\documentclass[letterpaper]{article} % DO NOT CHANGE THIS
\usepackage{aaai2026}  % DO NOT CHANGE THIS
\usepackage{times}  % DO NOT CHANGE THIS
\usepackage{helvet}  % DO NOT CHANGE THIS
\usepackage{courier}  % DO NOT CHANGE THIS
\usepackage[hyphens]{url}  % DO NOT CHANGE THIS
\usepackage{graphicx} % DO NOT CHANGE THIS
\usepackage{natbib}  % DO NOT CHANGE THIS AND DO NOT ADD ANY OPTIONS TO IT
\usepackage{caption} % DO NOT CHANGE THIS AND DO NOT ADD ANY OPTIONS TO IT
\usepackage{algorithm}
\usepackage{algorithmic}
\usepackage{amsmath}

\usepackage{newfloat}
\usepackage{listings}

\usepackage{multirow}
\usepackage{xcolor}
\usepackage{booktabs}

\DeclareCaptionStyle{ruled}{labelfont=normalfont,labelsep=colon,strut=off} % DO NOT CHANGE THIS
\floatstyle{ruled}
\newfloat{listing}{tb}{lst}{}
\floatname{listing}{Listing}
\title{EventHallusion: Diagnosing Event Hallucinations in Video LLMs}
\author{
    Jiacheng Zhang\textsuperscript{\rm 1,\rm 2}\thanks{Equal contribution.}, 
    Yang Jiao\textsuperscript{\rm 1,\rm 2}\footnotemark[1], 
    Shaoxiang Chen\textsuperscript{\rm 5},
    Na Zhao\textsuperscript{\rm 3},
    Zhiyu Tan\textsuperscript{\rm 4},
    Hao Li\textsuperscript{\rm 4},
    Xingjun Ma\textsuperscript{\rm 1,\rm 2}
    Jingjing Chen\textsuperscript{\rm 1,\rm 2}
}
\affiliations{
    \textsuperscript{\rm 1} Shanghai Key Lab of Intelligent Information Processing, School of CS, Fudan University\\
    \textsuperscript{\rm 2} Shanghai Collaborative Innovation Center on Intelligent Visual Computing\\
    \textsuperscript{\rm 3} Singapore University of Technology and Design\\
    \textsuperscript{\rm 4} Shanghai Academy of Artificial Intelligence for Science\\
    \textsuperscript{\rm 5} Meituan\\
    \{jiachengzhang22, yjiao23\}@m.fudan.edu.cn, 
    \{sxchen13, chenjingjing\}@fudan.edu.cn, 
    na\_zhao@sutd.edu.sg
}

\usepackage{bibentry}
\begin{document}
\maketitle

\begin{abstract}
Recently, Multimodal Large Language Models (MLLMs) have made significant progress in the video comprehension field. Despite remarkable content reasoning and instruction following capabilities they demonstrated, the hallucination problem of these VideoLLMs is less explored compared with its counterpart in the image domain. To mitigate this gap, we propose \textbf{EventHallusion}, a novel benchmark that focuses on assessing the VideoLLMs' hallucination toward \textbf{event}, the crux of video analysis. 
From a hallucination attribution perspective, our EventHallusion benchmark is curated to assess a VideoLLM's susceptibility toward language priors and vision-language biases.
On the other hand, we also propose a simple yet effective method, called Temporal Contrastive Decoding (TCD), to tackle the hallucination problems of VideoLLMs. The proposed TCD method rectifies the model's bias toward its priors during the decoding stage by comparing the original video with a modified version, in which temporal cues are made ambiguous. Through the comprehensive evaluation of eight open-source and three closed-source VideoLLMs on the proposed EventHallusion benchmark, we observe that the open-source models suffer significantly from hallucination problems, whereas the closed-source ones perform markedly better. By further equipping open-source VideoLLMs with the proposed TCD approach, evident performance improvements are achieved across most metrics in the EventHallusion benchmark. Our codes and benchmark data are available at https://github.com/Stevetich/EventHallusion.
\end{abstract}    
\begin{figure*}[t]
    \centering
    \includegraphics[width=1.\textwidth]{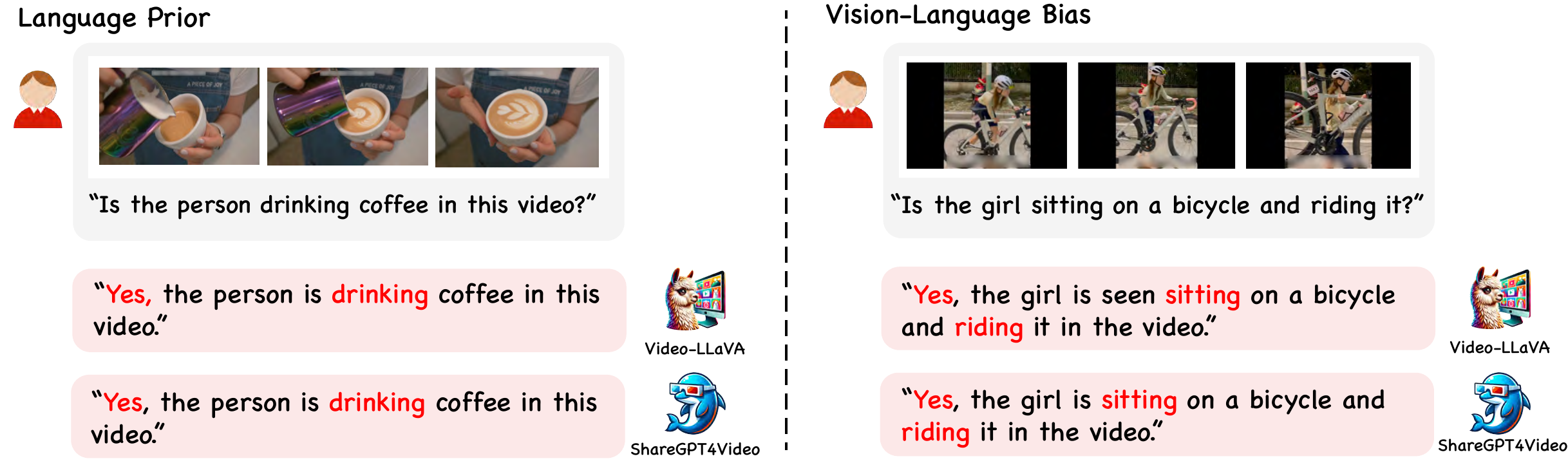}
    \caption{Illustration of two types of underlying reasons that may lead to existing VideoLLMs' hallucinations. We use red to indicate the hallucinations in answers generated by existing state-of-the-art VideoLLMs, namely Video-LLaVA and ShareGPT4Video.}
    \label{fig:intro}
    \vspace{-4mm}
\end{figure*}

\section{Introduction}
\label{sec:intro}

The field of deep learning has witnessed the extraordinary advancement of Large Language Models (LLMs)~\cite{brown2020language, touvron2023llama, dettmers2024qlora} and Multimodal Large Language Models (MLLMs)~\cite{alayrac2022flamingo, liu2024visual, bai2023qwen, jiao2024lumen, jiao2024rode}. Benefiting from dramatic semantic reasoning and instruction-following capabilities, LLMs and MLLMs have become versatile assistants across multiple disciplines ~\cite{jiao2024rode, yin2023foodlmm}. Despite their great success, they are still trapped in the hallucination problem, where models generate plausible but factually incorrect responses~\cite{leng2024vcd}, potentially threatening reliability and security in applying them in real-world scenarios~\cite{qian2024nuscenes,jiao2022more}. To systematically evaluate hallucinations of ImageLLMs, a thread of benchmarks has been developed~\cite{pope, fu2023mme, liu2023mmbench}. Specifically, POPE~\cite{pope} investigates the hallucination by mining the frequently occurring objects and their combinations. MME~\cite{fu2023mme} and MMBench~\cite{liu2023mmbench} evaluate hallucinations of the existence, attributes, and relations of objects.

% 简要写两句对于这几个benchmark的介绍

% \red{However, these benchmarks primarily assess hallucinations within static scenes, such as those involving objects, attributes, and inter-object relations, and therefore do not effectively address the needs of emerging VideoLLMs ~\cite{ma2023vista, llamavid}.} 
% However, when confronted with aforementioned contents within static scenes, existing VideoLLMs perform well due to the knowledge transferred from their foundation ImageLLMs.
% However, establishing upon the foundation of ImageLLMs, VideoLLMs perform well when comprehending the aforementioned contents in the single-frame image~\cite{wang2024videohallucer}, but significantly fall short in understanding the events (i.e, actions taken by the objects as defined in \red{ref}) happening in the whole video, as proved in Tab.\textcolor{red}{xxx}.
In fact, based on the foundation of ImageLLMs, VideoLLMs have shown significant proficiency in understanding the content of single-frame images~\cite{wang2024videohallucer}.
However, their effectiveness diminishes when tasked with comprehending the event, namely the actions performed by humans or objects spanning the entire video~\cite{du2024towards,zelnik2001event, krishna2017dense}. This limitation is further proved by our results presented in Fig.\ref{fig:obj_event}.
To further evaluate the event-oriented hallucinations of VideoLLMs, two recent benchmarks~\cite{wang2024videohallucer,guan2024hallusionbench} develop distinct event comprehension tasks. Specifically, HallusionBench~\cite{guan2024hallusionbench} reverses all frames within a video, subsequently querying the model to determine whether the original event occurs in the altered video. 
VideoHallucer~\cite{wang2024videohallucer} selects long videos containing multiple events and then queries the model to evaluate their chronological orders.
Although advancing the evaluation of event hallucination, they still suffer from the following limitations. 
On the one hand, the frame-reversed video in HallusionBench can deviate from the distribution of natural videos, as not all events remain meaningful when played in reverse. On the other hand, merely judging the chronological order of events given in the questions fails to probe the model's authentic ability to comprehend the complex events\footnote{A more detailed analysis of event-related VQA samples in HallusionBench~\cite{guan2024hallusionbench} and VideoHallucer~\cite{wang2024videohallucer} is provided in the supplementary materials due to page limitations.}.

\label{sec:intro}
% When delving into VideoLLMs' event-related hallucinations, we identify two primary causes rooted in the development processes of prevalent VideoLLMs, namely language prior and vision-language bias.
% Firstly, building upon the foundation of LLMs, VideoLLMs inherently inherit their language priors~\cite{agrawal2016analyzing,agarwal2020towards}, which manifests as an overconfidence in their language knowledge. 
% As shown in the left column of Fig.\ref{fig:intro}, both Video-LLaVA\cite{videollava} and ShareGPT4Video\cite{chen2024sharegpt4video} are easily misled by the frequently occurring but misleading event \textit{``person drink coffee"} asked in the question.
% Moreover, due to the high cost associated with scaling up high-quality video-language data, prevalent VideoLLMs typically acquire multimodal knowledge from large-scale image-language datasets. 
% However, in these datasets, specious event-related captions often arise based solely on a single image, which inevitably introduces spurious scene-event correlations. 
% As shown in the right column of Fig.\ref{fig:intro}, both Video-LLaVA\cite{videollava} and ShareGPT4Video\cite{chen2024sharegpt4video} intuitively make the conclusion of \textit{``cats are playing"} when perceiving two cats within the video, without respecting the truth.

When delving into event-related hallucinations in VideoLLMs, we observe that such errors originate from a consistent issue across different training stages: imbalanced training data distributions, which inject prior knowledge and bias into the model. Initially, building upon LLM backbones, VideoLLMs memorize events with higher frequencies purely through textual statistics, thereby forming strong language-driven priors~\cite{agrawal2016analyzing,agarwal2020towards}. For instance, the commonly described event \textit{``person drinks coffee''} frequently appears in textual corpora, establishing a misleading prior that persists regardless of visual evidence (Fig.\ref{fig:intro}, left). 
Moreover, due to the high cost associated with scaling up high-quality video-language data, prevalent VideoLLMs typically acquire multimodal knowledge from large-scale image-language datasets. 
However, similar imbalances surface during the vision-language alignment stage, where frequently co-occurring visual-textual pairs reinforce spurious object-event associations. As shown in the right column of Fig.\ref{fig:intro}, both Video-LLaVA~\cite{videollava} and ShareGPT4Video~\cite{chen2024sharegpt4video} intuitively conclude \textit{``people are riding the bicycle''} simply from the presence of a person and a bicycle in the scene, disregarding the actual video context.

\begin{figure*}[t]
    \centering
    \includegraphics[width=0.8\textwidth]{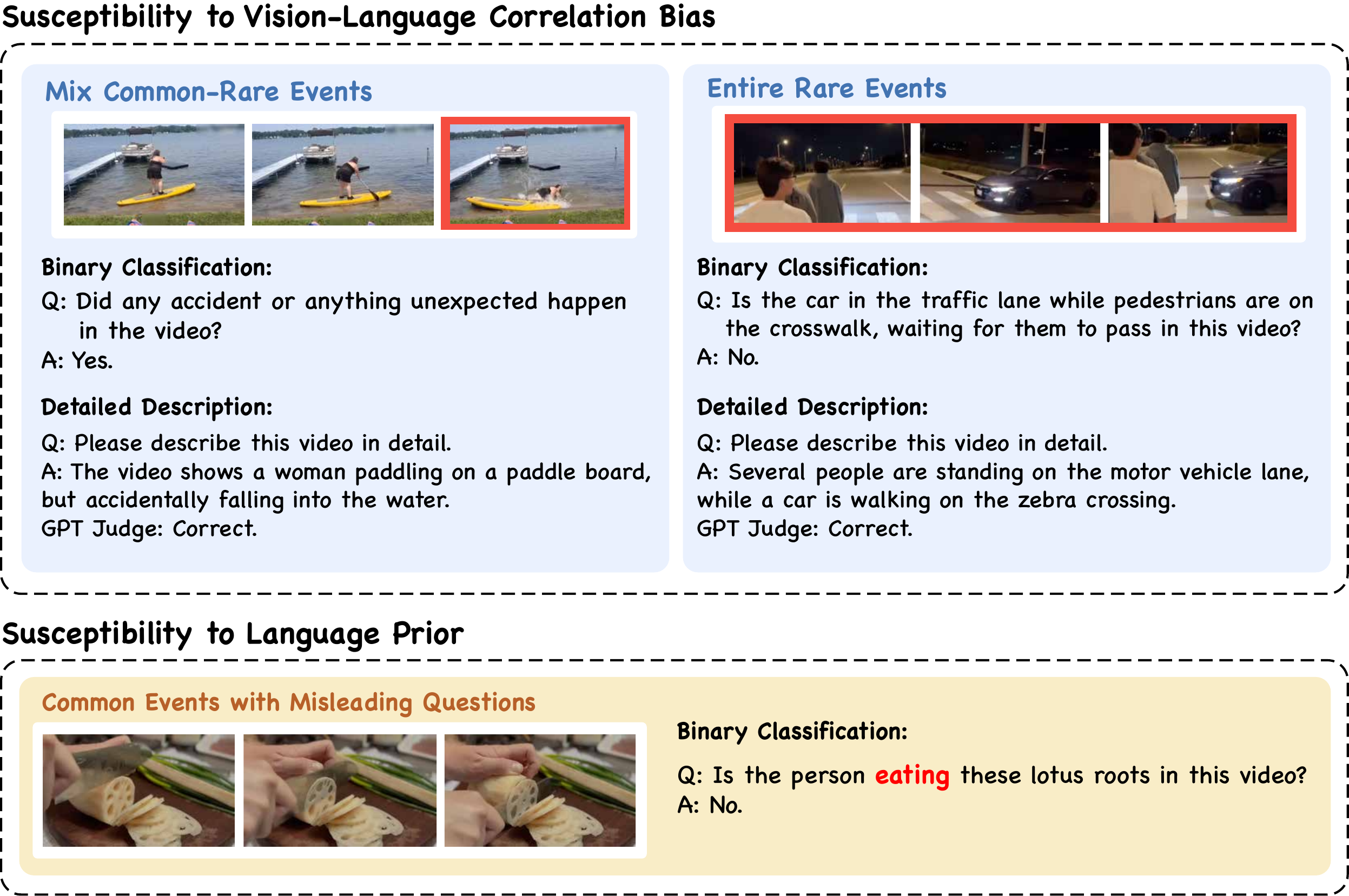}
    \caption{Illustration of examples in our EventHallusion benchmark. Questions are basically divided into two categories: \textit{Susceptibility to Language Prior} and \textit{Susceptibility to Vision-Language Correlation Bias}. The contents that induces VideoLLMs to generate event-related hallucinations are marked by red.}
    \label{fig:intro_2}
    \vspace{-2mm}
\end{figure*}

Toward this end, in this paper, we propose EventHallusion, a novel benchmark that systematically evaluates event-related hallucinations of state-of-the-art VideoLLMs. From a hallucination attribution perspective, our EventHallusion is curated to assess the susceptibility of a VideoLLM to language priors and vision-language correlation biases, respectively. Specifically, our benchmark consists of two basic categories as shown in Fig.\ref{fig:intro_2}: (1) \emph{\textbf{Susceptibility to language priors}}: In this category, all questions are annotated with a misleading event designed to prompt the model to make decisions based on its language priors. Within such events, while objects align with those depicted in the video, their actions are deliberately misrepresented.
% As shown in Fig.\red{XX}, 
(2) \emph{\textbf{Susceptibility to vision-language correlation biases}}: In this category, all videos are meticulously selected to ensure that the included events are infrequent yet plausible. When presented with these infrequent events, the model can easily be misled to interpret the event as one that frequently co-occurs with the objects or scenes depicted in the video, based on its memory.
It is worth mentioning that through comprehensive evaluation, we find that existing open-source VideoLLMs suffer from severe event hallucination issues on our benchmark, which could offer promising avenues for enhancing the VideoLLM's capabilities in the future research.

% Moreover, inspired by a hallucination elimination approach for image LLMs (i.e., VCD~\cite{leng2024vcd}), 
Additionally, we propose a simple yet effective method called Temporal Contrastive Decoding (TCD) to reduce the event-related hallucinations of existing VideoLLMs in a training-free manner.
The core motivation of our TCD approach is to first elicit the spurious priors and then eliminate their adverse effects before making the final decisions.
Technically, for each given video, we first construct its counterpart by ambiguating the temporal cues while preserving the spatial information. This counterpart disentangles the event-unrelated elements that could potentially trigger spurious prior assumptions from the original video. Therefore, we feed both the original video and its counterpart into a VideoLLM and adaptively adjust the prediction distribution of the former using the latter, leading to more reliable final judgments. As a plug-and-play approach, our TCD can consistently boost existing VideoLLM's performances on the proposed EventHallusion as shown in Tab.\ref{tab:tcd}.
% As previously analyzed, VideoLLMs can be easily influenced by their prior knowledge. Therefore, our TCD is designed to mitigate such biases and enhance the model's objective judgments in event comprehension. 
% Without any additional training budgets, our TCD rectifies the model's bias to its prior in an adaptive manner during the decoding stage. 
% Technically, for each given video, we first apply the temporal downsampling operation to create a modified counterpart. Both the original video and its modified counterpart are then input into the VideoLLMs. During the decoding stage, the logits prediction of the original video is adjusted by the logits prediction of the counterpart. The rectified logits are then used to sample the next token and continue the decoding process.
% % similar to the approach in VCD~\cite{leng2024vcd}.
% Within the proposed TCD, the modified counterpart disrupts the temporal cues while preserving spatial object and scene information, making its predicted logits more susceptible to bias from the model's event-related prior knowledge. Subtracting these amplified priors in the logits space during the decoding process helps guide the model to make more reliable judgments. As a plug-and-play approach, our TCD can consistently boost existing VideoLLM's performances on the proposed EvenHallusion as shown in Tab.\ref{tab:tcd}.

In summary, our contributions are threefold: (1) We propose EventHallusion, a novel video QA benchmark focused on assessing event-related hallucinations in VideoLLMs; (2) We introduce Temporal Contrastive Decoding (TCD), a simple yet effective approach that can reduce hallucinations across a wide range of VideoLLMs in a plug-and-play manner; (3) We also conduct comprehensive ablation studies to provide more in-depth analysis.

\section{Related Works}
\label{sec:related}

\textbf{Video Large Language Models.}
Large Language Models (LLMs) have achieved remarkable success in traditional NLP tasks and exhibit strong generalization to user-defined instructions, enabling them to act as versatile assistants. Building on this foundation, Multimodal Language Models (MLLMs) extend LLMs with visual inputs, facilitating unified text-vision understanding. To further incorporate video comprehension and temporality, video-supported MLLMs have emerged. Methods such as Video-ChatGPT~\cite{maaz2023videochatgpt} apply spatial-temporal pooling, while Video-LLaMA~\cite{zhang2023videollama} introduces a Video Q-Former for frame aggregation. Vista-LLaMA~\cite{ma2023vista} enhances temporal modeling via a temporal Q-Former projector, and LLaMA-VID~\cite{llamavid} proposes a dual-token paradigm to represent both context and content per frame. Although these approaches perform well on benchmarks, they still struggle with hallucinations in temporal understanding.
% In the natural language processing field, Large Language Models (LLMs) have demonstrated astounding performance in solving traditional tasks. What's more, LLMs also exhibit extraordinary generalization in handling self-defined tasks described by user's instructions, making them capable of playing the role of versatile assistants. Benefiting from such abilities, Multimodal Language Models (MLLMs) have emerged by introducing visual inputs into LLMs, promoting the unified comprehension of both textual and visual content. 

% Video-supported MLLMs are further proposed since the established vision-text understanding ability of image MLLMs. To enable MLLMs to comprehend video content and the concept of temporality, numerous methods have been put forward. Video-ChatGPT \cite{maaz2023videochatgpt} utilizes spatial and temporal pooling operations to conduct video modeling, while Video-LLaMA \cite{zhang2023videollama} introduces a Video Q-Former to aggregate frame-level representations. Vista-LLaMA \cite{ma2023vista} preserves the equal distance between all visual tokens and any language tokens and proposes the temporal Q-Former projector to further enhance the representations of the video and the modeling of temporality. LLaMA-VID \cite{llamavid} proposes a dual-token paradigm, using a context token and a content token to represent each frame, which effectively supports the comprehension of long videos and maintains temporal information. However, despite their excellent performance on existing benchmarks, these methods still suffer the hallucination problem in temporality.

\noindent\textbf{Video Evaluation Benchmarks.}
Despite the rapid progress of MLLMs, a comprehensive evaluation of their capabilities remains essential. For image MLLMs, various benchmarks have been developed to assess perception, reasoning, and cognition~\cite{fu2023mme, liu2023mmbench, li2024eyes}. With the emergence of VideoLLMs, benchmarks like TempCompass~\cite{liu2024tempcompass} and FunQA~\cite{xie2023funqa} have been proposed to evaluate temporal perception and complex video comprehension.

Despite previous works sharing similarities with ours, several critical differences persist.
First, our work specifically investigates event-related hallucinations in VideoLLMs, attributing their occurrence to language priors and vision-language bias inherited during pretraining. Building on this insight, we curate a benchmark with event-prone videos to facilitate targeted evaluation. Furthermore, we introduce Temporal Contrastive Decoding (TCD), a simple yet effective method to alleviate event hallucinations, demonstrating general improvements across multiple VideoLLMs.
% Despite the extraordinary advancement of MLLMs, it is of great significance to conduct quantitative evaluations of existing models. To thoroughly evaluate the performance of image MLLMs, a wide scope of efforts has been made~\cite{fu2023mme, liu2023mmbench, li2024eyes}. Their evaluation comprises various aspects, including perception, reasoning, and cognition. With the development of VideoLLMs, several benchmarks have also been proposed to assess video comprehension capacities. Tempcompass \cite{liu2024tempcompass} focuses on the thorough evaluation of the temporal perception ability of VideoLLMs, including action, speed, direction, attribute change, and event order. FunQA \cite{xie2023funqa} proposes utilizing humorous, creative, and magic videos to assess the in-depth comprehension ability of existing VideoLLMs. 

% Despite previous works sharing similarities with ours, there exist several critical differences. 
% First, we focus on exploring and evaluating the hallucination phenomenon in the event aspects of existing VideoLLMs. Furthermore, we emphasize and validate that the event hallucination phenomenon of video LLMs is influenced by language prior. 
% Second, based on this conclusion, we collect videos that are inclined to cause event hallucinations in VideoLLMs and construct the benchmark for evaluation.
% Third, we propose a Temporal Contrastive Decoding (TCD) to alleviate the event hallucination, which exhibits effectiveness in multiple VideoLLMs.

\noindent\textbf{Hallucination in MLLMs.}
Hallucination is a phenomenon widely observed in MLLMs, arising from factors such as imbalanced training data, weak attention to visual content, and other possible reasons. To mitigate hallucinations in image MLLMs, several strategies have been proposed. Woodpecker~\cite{yin2023woodpecker} introduces a post-hoc correction mechanism leveraging additional visual cues, while Volcano~\cite{lee2023volcano} adopts a critique-revise-decide framework for iterative self-revision. From a data perspective, LRV-Instruction~\cite{lrv_instruction} and HalluciDoctor~\cite{yu2024hallucidoctor} aim to alleviate hallucinations by constructing more balanced, high-quality instruction tuning datasets.

Another line of work addresses hallucinations during inference, where contrastive decoding (CD) has shown effectiveness without requiring additional training. VCD~\cite{leng2024vcd} mitigates statistical bias and language priors by contrasting original and distorted images during decoding. However, existing CD methods mainly target ImageLLMs, with limited exploration in the video domain. Motivated by this gap, we propose Temporal Contrastive Decoding (TCD) to reduce event hallucinations in VideoLLMs.
% Hallucination is a phenomenon widely observed in MLLMs. The occurrence of hallucinations is attributed to various factors, including imbalanced training data distribution, weak attention to visual content, and other possible reasons. To alleviate the impact of hallucinations in image MLLMs, several methods have been proposed. Woodpecker \cite{yin2023woodpecker} adopts a post-hoc correction mechanism that utilizes additional extracted visual cues to rectify previously generated responses, while Volcano \cite{lee2023volcano} proposes a critique-revise-decide process to self-revise their initial generated answers. Besides, LRV-Instruction \cite{lrv_instruction} and HalluciDoctor \cite{yu2024hallucidoctor} propose to mitigate this problem from the data perspective to provide more balanced, high-quality instruction tuning datasets.

% Some other works propose to alleviate hallucination during the inference phase, in which the contrastive decoding (CD) method has been widely adopted due to its excellent performance and the absence of additional training. VCD \cite{leng2024vcd} employs the original and distorted images for comparison in contrastive decoding, thereby reducing the impact of statistical bias and language priors. Despite the extensive exploration of adopting CD in ImageLLMs, there is still a lack of effort in employing CD to handle hallucinations in VideoLLMs. Inspired by this, we propose Temporal Contrastive Decoding (TCD) to mitigate event hallucinations in VideoLLMs.

\begin{figure}[!t]
    \centering
    \includegraphics[width=0.45\textwidth]{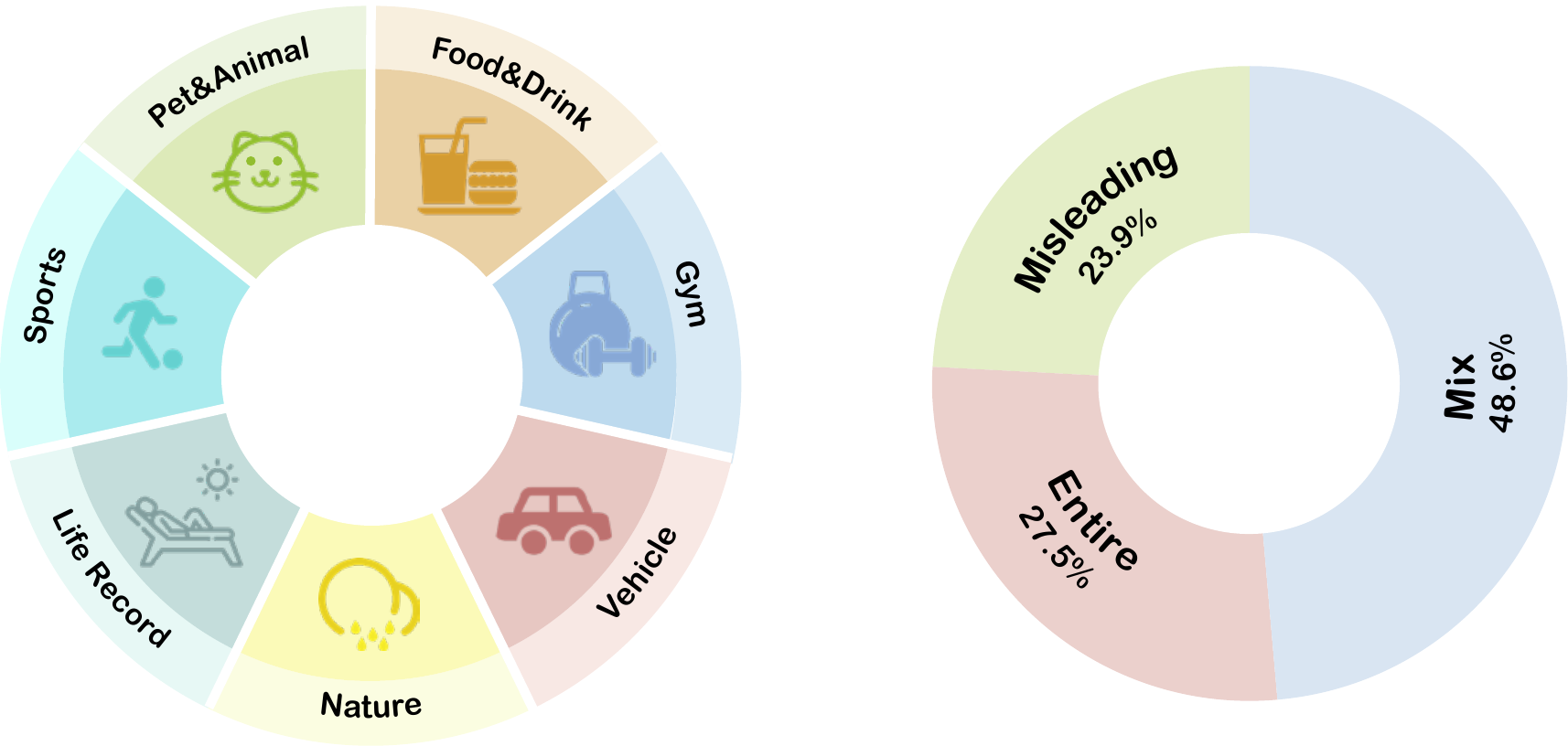}
    \caption{Video category and questions distribution. ``Entire" implies that the entire video depicts the rare event. ``Mix" implies that common and rare events interleavedly occur in the video. ``Misleading" implies the video with the common event, but with its question customized to include misleading event demonstration.}
    \label{fig:metrics_1}
    \vspace{-0mm}
\end{figure}

\section{EventHallusion Benchmark}
In this section, we first elaborate on the data curation process of two basic categories of our EventHallusion benchmark. Afterward, we demonstrate the distribution statistics of the whole EventHallusion benchmark. Finally, we
introduce the specific evaluation protocol of different question types.

\begin{figure}[!t]
    \centering
    \includegraphics[width=0.45\textwidth]{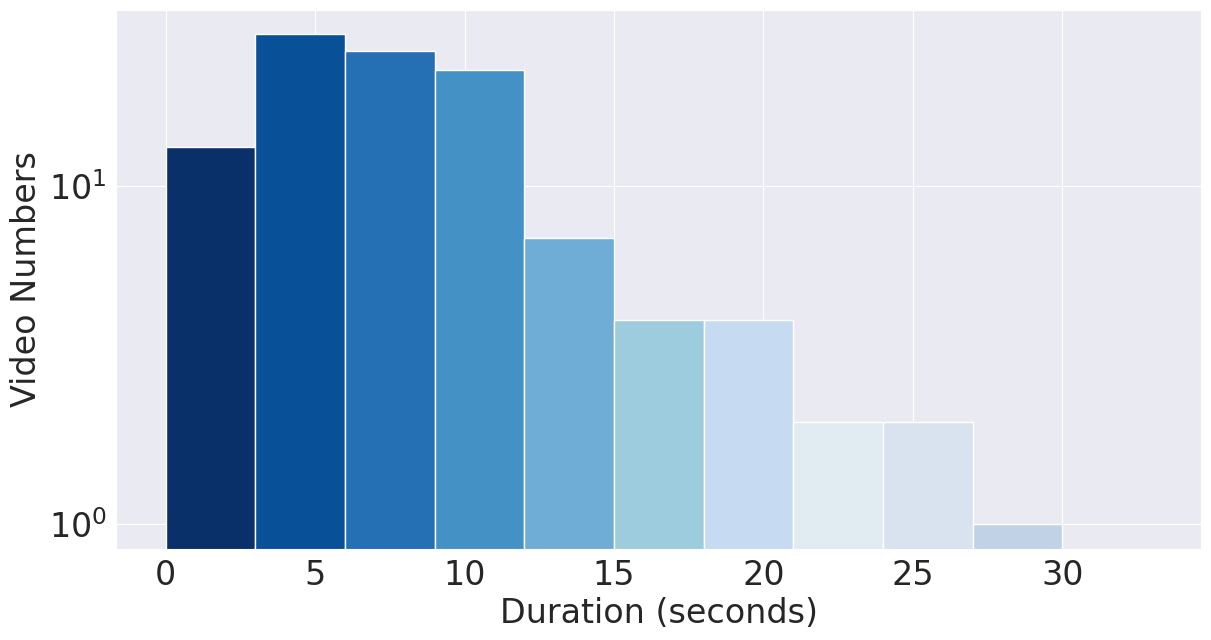}
    \caption{The distribution of video length within our EventHallusion benchmark.}
    \label{fig:metrics_2}
    \vspace{-3mm}
\end{figure}

\subsection{Data Curation}
\label{sec:data_curation}
\noindent\textbf{Susceptibility to language priors.} In this category, we first collect videos describing the events that occur in daily life from the internet. To avoid information leakage, we blur the text captions printed on videos, thereby encouraging models to analyze events solely based on visual content.  
However, when annotating the questions, we explicitly include another frequently occurring event that is inconsistent with the video content to trigger the language priors within the LLM. With these ``misleading triggers", we formulate questions as a binary choice problem as shown below:
\begin{center}
    \textit{``Did the [subject] [event] happen in the video?''}
\end{center}
Here, the combination of \textit{[subject]} and \textit{[event]} correspond to the ``misleading triggers" aforementioned.
In Fig.\ref{fig:intro_2}, we refer to this type of sample as \textit{``Common Events with Misleading Questions"}.

\begin{figure*}[t]
    \centering
    \includegraphics[width=1.\textwidth]{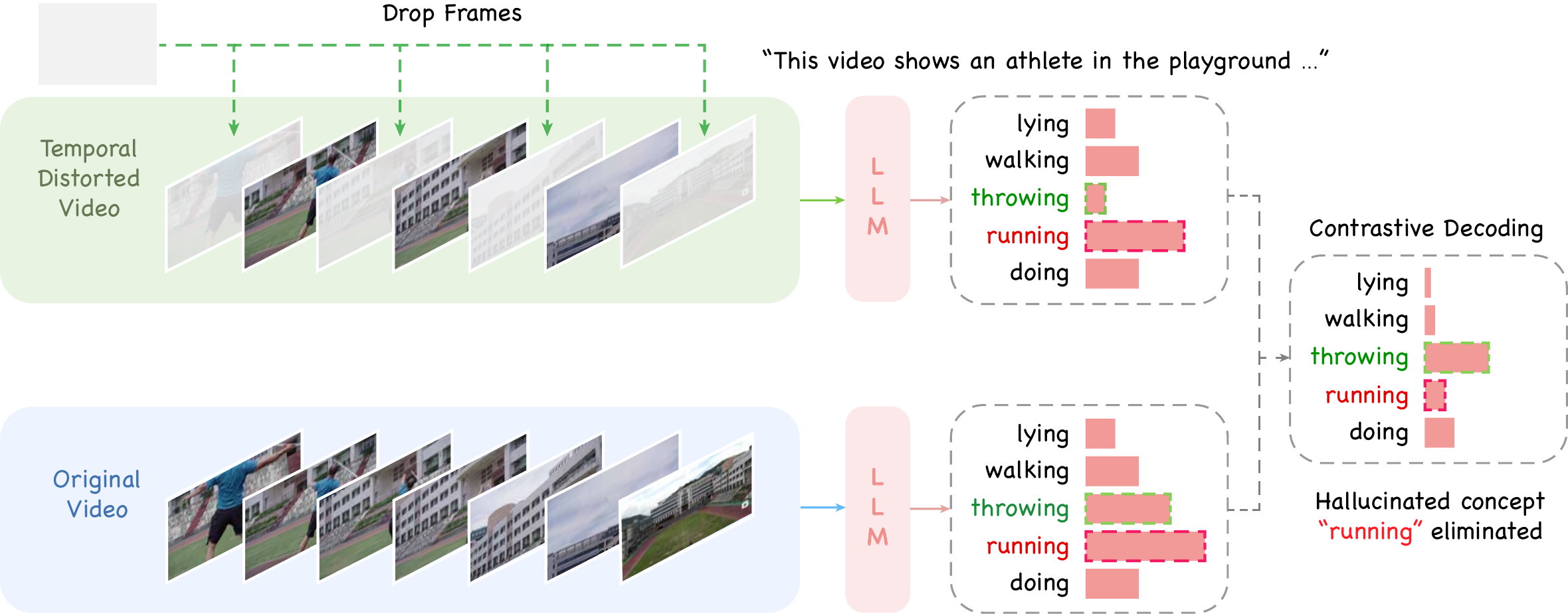}
    \caption{Framework of Temporal Contrastive Decoding. We construct the temporal modeling distorted video as the comparison group with fewer frames extracted and then contrast their logits to mitigate the event hallucination.}
    \label{fig:framework}
    \vspace{-2mm}
\end{figure*}

\noindent\textbf{Susceptibility to vision-language correlation biases.} 
In this category, the primary challenge lies in sourcing the appropriate video data. Since existing publicly available video datasets (e.g., ActivityNet~\cite{caba2015activitynet}, \cite{das2013thousand}, \cite{soomro2012ucf101}) collect videos from daily life, the events they include cater to the vision-language correlation biases stored in VideoLLMs, resulting in high recognition precision. Therefore, instead of refactoring existing video datasets, we meticulously collect videos from prevalent Internet video platforms and filter them with rigorous cross-validation. Specifically, we begin by manually collecting videos that depict infrequent events based on human consensus. To minimize personal bias, we engage three annotators to assess whether the events in these selected videos are indeed uncommon. Only videos that receive unanimous positive judgments from all annotators are retained as testing samples. Similar to the first category, we also employ the blurring operation on videos to prevent information leakage. We further divide these videos into two groups: (1) Videos where infrequent events occur throughout the entire duration, referred to as ``\textit{Entire Rare Events}" in Fig.\ref{fig:intro_2}. (2) Videos containing a mix of both frequent and infrequent events, referred to as ``\textit{Mix Common-Rare Events}" in Fig.\ref{fig:intro_2}. 
% \blue{(TODO: Discuss Mix is more challenging than Entire if experimental results are supported ...)}

For constructing questions, we design both binary and open-ended types for comprehensive evaluation. Specifically, the binary questions ask the model whether a plausible but incorrect common event occurs in the given video. The open-ended questions, on the other hand, simply prompt the model to freely describe the video content. Formally, binary and open-ended questions for the \textit{``Entire"} category can be formulated as presented below:
\begin{center}
    Binary: \textit{``Did/Is the [subject] [event] in the video?''}, \\
    Open-Ended: \textit{``Please describe this video in detail.''},
\end{center}
For the \textit{``Mix"} category, since both common and rare events can happen in the video, we reformulate the binary questions as below: 
\begin{center}
    \textit{``Did the [subject] [event] in the entire video?''}, \\
    \textit{``Did any accident or anything unexpected happen in the video?''},
\end{center}
For the ground-truth answers to all types of questions, we meticulously verify them to ensure their correctness.

\noindent\textbf{Dataset distribution statistics.}
In total, our EventHallusion benchmark comprises 400 videos and 711 questions.  
To ensure comprehensive coverage across diverse domains, these videos span seven scenarios, including \textit{Pet\&Animal, Sports Competition, Food\&Drink, Gym Exercises, Vehicle, Life Record, Nature}, as shown in Fig.\ref{fig:metrics_1}. The distribution of video lengths in our benchmark is presented in Fig.\ref{fig:metrics_2}, showing that most videos are relatively short. This is because longer videos often involve significant scene fluctuations, which introduce additional challenges for VideoLLMs beyond the uni-modal and multi-modal biases we aim to examine in this paper.

\subsection{Evaluation Protocol}
\noindent\textbf{Binary question answering.} Following previous benchmarks~\cite{wang2024videohallucer,pope}, we compare the first word of the model's answer with the ground-truth answer to determine whether the model answered the question correctly. This approach is justified by the VideoLLM's exceptional instruction-following capability, which typically enables it to answer binary choice questions with \textit{``yes"} or \textit{``no"} at the beginning. We calculate the accuracy as the evaluation metric for binary questions.

\noindent\textbf{Open-ended question answering.} For automatic evaluation, we feed both generated and ground-truth answers into GPT-4o\cite{hurst2024gpt4o}, using it as a knowledgeable evaluator to determine whether the model answers the question correctly following prior benchmarks \cite{liu2024tempcompass}. We also use accuracy as the evaluation metric. Detailed templates of prompting GPT-4o are provided in the supplementary materials due to page limitation. 

\begin{table*}[ht]
\centering
\scalebox{0.9}{
\begin{tabular}{@{}lccccclcc@{}}
\toprule
\multicolumn{1}{c}{\multirow{2}{*}{\textbf{Models}}} &
  \multicolumn{2}{c}{\textbf{Entire}} &
  \multicolumn{2}{c}{\textbf{Mix}} &
  \multicolumn{2}{c}{\textbf{Misleading}} &
  \multicolumn{2}{c}{\textbf{Overall}} \\ \cmidrule(l){2-9} 
\multicolumn{1}{c}{} &
  Binary &
  Desc. &
  Binary &
  Desc. &
  \multicolumn{2}{c}{Binary} &
  Binary &
  Desc. \\ \midrule
\textit{Open-source Models} &
  \multicolumn{1}{l}{} &
  \multicolumn{1}{l}{} &
  \multicolumn{1}{l}{} &
  \multicolumn{1}{l}{} &
  \multicolumn{2}{l}{} &
  \multicolumn{1}{l}{} &
  \multicolumn{1}{l}{} \\ \midrule
\multicolumn{1}{l|}{VideoChatGPT \cite{maaz2023videochatgpt}} &
  14.91 &
  5.50 &
  56.99 &
  3.63 &
  \multicolumn{2}{c|}{21.57} &
  36.43 &
  4.30 \\
\multicolumn{1}{l|}{VideoChat2 \cite{li2024mvbench}} &
  16.67 &
  4.59 &
  47.67 &
  1.55 &
  \multicolumn{2}{c|}{22.55} &
  32.76 &
  2.64 \\
\multicolumn{1}{l|}{PLLaVA \cite{xu2024pllava}} &
  45.61 &
  16.51 &
  58.55 &
  3.11 &
  \multicolumn{2}{c|}{81.37} &
  60.64 &
  6.05 \\
\multicolumn{1}{l|}{LLaMA-VID \cite{llamavid}} &
  30.70 &
  16.51 &
  \textbf{73.58} &
  7.77 &
  \multicolumn{2}{c|}{43.14} &
  54.03 &
  10.90 \\
\multicolumn{1}{l|}{LLaVA-NeXT-Video \cite{li2024llava}} &
  49.12 &
  9.17 &
  70.47 &
  4.66 &
  \multicolumn{2}{c|}{71.57} &
  \textbf{64.79} &
  6.29 \\
\multicolumn{1}{l|}{VILA \cite{lin2024vila}} &
  \textbf{53.51} &
  \textbf{20.18} &
  58.55 &
  \textbf{17.62} &
  \multicolumn{2}{c|}{\textbf{83.33}} &
  63.33 &
  \textbf{18.54} \\
\multicolumn{1}{l|}{ShareGPT4Video \cite{chen2024sharegpt4video}} &
  11.40 &
  0.00 &
  67.88 &
  5.18 &
  \multicolumn{2}{c|}{6.86} &
  49.14 &
  9.82 \\
\multicolumn{1}{l|}{Video-LLaVA \cite{videollava}} &
  30.70 &
  8.26 &
  57.51 &
  7.25 &
  \multicolumn{2}{c|}{41.18} &
  45.97 &
  7.62 \\ \midrule
\textit{Closed-source Models} &
  \multicolumn{1}{l}{} &
  \multicolumn{1}{l}{} &
  \multicolumn{1}{l}{} &
  \multicolumn{1}{l}{} &
  \multicolumn{2}{l}{} &
  \multicolumn{1}{l}{} &
  \multicolumn{1}{l}{} \\ \midrule
\multicolumn{1}{l|}{GPT-4V \cite{achiam2023gpt}} &
  57.89 &
  11.01 &
  79.27 &
  16.58 &
  \multicolumn{2}{c|}{92.16} &
  76.53 &
  14.57 \\
\multicolumn{1}{l|}{Gemini-1.5-Pro \cite{team2023gemini}} &
  61.40 &
  49.54 &
  83.42 &
  39.90 &
  \multicolumn{2}{c|}{96.08} &
  80.44 &
  43.38 \\
\multicolumn{1}{l|}{GPT-4o \cite{hurst2024gpt4o}} &
  \textbf{83.33} &
  \textbf{58.72} &
  \textbf{92.75} &
  \textbf{41.97} &
  \multicolumn{2}{c|}{\textbf{100.00}} &
  \textbf{91.93} &
  \textbf{48.01} \\ \bottomrule
\end{tabular}}
\caption{\textbf{Performance comparison of VideoLLMs on EventHallusion.} We exhibit the \textbf{accuracy} of both open-source and closed-source models on the binary classification task and the detailed description matching task on our three types of videos, respectively. We highlight the best-performing results for both closed-source and open-source VideoLLMs in bold. In addition, we show the overall success rate of each model on our binary classification task that answers "yes" or "no" in the first word.}
\label{tab:main_table}
\end{table*}

\section{Temporal Contrastive Decoding}
To mitigate the aforementioned hallucinations in VideoLLMs, the most straightforward solution is to enhance finetuning with additional debiased data. However, this approach is highly resource-intensive, as it requires extensive efforts to collect such videos and annotate corresponding QA pairs. Therefore, we take inspiration from~\cite{leng2024vcd} and propose our Temporal Contrastive Decoding (TCD) method, which alleviates event-oriented hallucinations in VideoLLMs by comparing the logits of the original video and its counterpart during the decoding stage.

% \red{Motivated by the previous work~\cite{leng2024vcd}, our TCD method aims to reduce event hallucinations by adjusting the distribution of predicted logits at each decoding step, using a contrastive counterpart that is more aligned with the model's inherent biases.} 
% As previously analyzed in Sec.\ref{sec:intro}, the major causes of event hallucination stem from the spurious language and vision-language priors. The core motivation behind our TCD approach is to first elicit these priors using a specifically designed contrastive counterpart of the original video, and then subtract them in an adaptive manner from the original responses to reveal the accurate responses.

For clarity, we first review the standard auto-regressive decoding process of a VideoLLM. Given the text instruction $x$ and the video $V$, the predicted logit $z_{t}^{ori}$ at the $t$\textit{-th} timestep can be formulated as:
\begin{equation}
z_{t}^{ori}=\operatorname{logit}_{\theta}\left(y_{t} \mid V, x, y_{<t}\right)
\label{eq:standard_AR}
\end{equation}
where $y_t$ is the token to be generated at time step $t$, and $y_{<t}$ denotes the generated token sequence before time step $t$. $y_t$ is based on $x$, $V$, and all previously generated tokens $y_{<t}$ by selecting the token with the highest probability.

\begin{figure}[ht]
    \centering
    \includegraphics[width=0.4\textwidth]{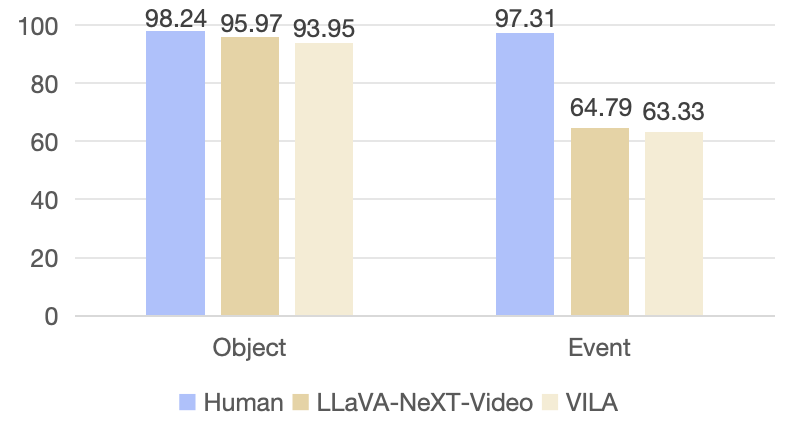}
    \vspace{-1mm}
    \caption{
Performance comparison of prevalent VideoLLMs and human evaluations on object-related and event-related content recognition accuracy (i.e., our official EventHallusion evaluation) in our EventHallusion benchmark.}
    \label{fig:obj_event}
    \vspace{-3mm}
\end{figure}

When constructing a contrastive counterpart, we chronologically downsample the original video as shown in the Fig.\ref{fig:framework}, aiming to weaken the temporal cues while emphasizing object cues. Therefore, the contrastive counterpart's probability prediction $z_{t}^{con}$ at timestep $t$ can be formulated as:
\begin{equation}
    z_{t}^{con}=\operatorname{logit}_{\theta}\left(y_{t} \mid S(V), x, y_{<t}\right)
\end{equation}
where $S(\cdot)$ represents the temporal downsampling operation. 
With the generated $z_{t}^{ori}$ and $z_{t}^{con}$, the modulated logit at the current timestep $t$ can be calculated as:
% \begin{equation}
%    p_{t} = \operatorname{softmax}\Big[(1+\alpha)z_{t}^{ori}-\alpha z_{t}^{con}\Big]
% \label{eq:tcd}
% \end{equation}
\begin{equation}
   z_{t} = (1+\alpha)z_{t}^{ori}-\alpha z_{t}^{con}
\label{eq:tcd}
\end{equation}

\noindent where $\alpha$ is the hyperparameter to control the contrastive strength during the decoding.
Since $z_{t}^{ori}$ and $z_{t}^{con}$ are not always exhibited contrastively, we eliminate the adverse effects of suppressing true positives by directly filtering 
implausible answers with a dynamic confidence threshold $t$:
\begin{equation}
    t = \beta*\mathrm{max} (z^{ori}_{t})
\end{equation}
The ultimate modulated probability $p_{t}$ can be formulated as:
% \begin{equation}
%     % p_{t}[k] = \begin{cases} 
%     % \operatorname{softmax}\Big(z_{t}[k]\Big) & \text{if } z_{t}[k] >= t, \\
%     % -\infty & \text{if } z_{t}[k] < t.
%     % \end{cases}
% \end{equation}
\begin{equation}
\begin{aligned}
\tilde{z}_t[k] &=
\begin{cases}
z_t[k], & \text{if } z_t[k] \ge t, \\
-\infty, & \text{otherwise},
\end{cases} \\
p_t &= \mathrm{softmax}(\tilde{z}_t).
\end{aligned}
\end{equation}
where $z_{t}[k]$ denotes the $k$-th dimension of $z_{t}$.

\begin{table}[t]
\centering
\scalebox{0.8}{
\begin{tabular}{@{}llcccc@{}}
\toprule
Model & Decoding & Entire         & Mix            & Mis.     & Overall        \\ \midrule
\multirow{2}{*}{LNV~\cite{li2024llava}} & Default & 49.12          & 70.47 & 71.57 & 64.79 \\
      & TCD      & \textbf{50.88} & \textbf{73.58} & \textbf{74.51} & \textbf{67.48} \\ \midrule
\multirow{2}{*}{VC2 \cite{li2024mvbench}}     & Default & 16.67          & 47.67 & 22.55 & 32.76 \\
      & TCD      & \textbf{17.54} & \textbf{49.74} & \textbf{25.49} & \textbf{34.72} \\ \midrule
\multirow{2}{*}{VILA \cite{lin2024vila}}             & Default & \textbf{53.51} & 58.55 & 83.33 & 63.33 \\
      & TCD      & 50.88          & \textbf{66.32} & \textbf{84.31} & \textbf{66.50} \\ \bottomrule
\end{tabular}}
\caption{Results of TCD on EventHallusion. \textit{Default} denotes the standard decoding strategy, while \textit{TCD} represents our Temporal Contrastive Decoding method. We report the \textbf{accuracy} of the binary classification task. The best results of each category are \textbf{bolded}. ``LNV" and ``VC2'' are short for ``LLaVA-NeXT-Video'' and ``VideoChat2'', respectively. ``Mis'' denotes the ``Misleading'' category.}
\label{tab:tcd}
\vspace{-3mm}
\end{table} 

\section{Experiments}
In this section, we first evaluate multiple prevalent VideoLLMs on our EventHallusion, including both open-source and closed-source MLLMs. Subsequently, we examine the effectiveness of our proposed TCD and also discuss its more specific design choices. 
% th EventHallusion with three VideoLLMs and demonstrate its effectiveness in improving the robustness of VideoLLMs against event hallucinations.

\subsection{Evaluation on EventHallusion}

\subsubsection{Performances of Human Evaluation}
Firstly, to evaluate the quality of our proposed EventHallusion benchmark, we ask three humans to answer the questions and average their resulting scores as the human evaluation results. 
Meanwhile, to validate the value of evaluating event-oriented hallucinations for VideoLLMs, we also extracted object contents from videos in our EventHallusion benchmark and tested them using both human evaluations and two prevalent VideoLLMs. The above results are shown in Fig.\ref{fig:obj_event}, where we have the following conclusions: (1) Although prevalent VideoLLMs perform poorly in EventHallusion (i.e., ``Event" part in Fig.\ref{fig:obj_event}), humans can correctly answer almost all questions. This demonstrates that the questions in our EventHallusion benchmark are clear enough to allow for accurate judgments. (2) When recognizing objects in videos, prevalent VideoLLMs can achieve human-level accuracies. However, they significantly lag behind humans when comprehending event-related content. This demonstrates that event-oriented hallucinations confuse VideoLLMs much more severely than object-related ones.

\subsubsection{Performances of Prevalent VideoLLMs}
\textbf{Evaluation Setup.}\quad We test 8 open-source Video and 3 closed-source VideoLLMs on our EventHallusion. Open-source VideoLLMs comprise VideoChatGPT \cite{maaz2023videochatgpt}, VideoChat2 \cite{li2024mvbench}, PLLaVA \cite{xu2024pllava}, LLaMA-VID \cite{llamavid}, LLaVA-NeXT-Video \cite{li2024llava}, VILA \cite{lin2024vila}, ShareGPT4Video \cite{chen2024sharegpt4video}, and Video-LLaVA \cite{videollava}. For closed-source VideoLLMs, we select the three most popular video-supported MLLMs, namely GPT-4V \cite{achiam2023gpt}, GPT-4o\cite{hurst2024gpt4o}, and Gemini-1.5-Pro \cite{team2023gemini}, to conduct the systematic comparison of the robustness between open-source and closed-source MLLMs.

For the inference setting of each VideoLLM, we follow their original setting, including conversation mode and other hyperparameters. For the number of frames, we sample at least 8 frames for each VideoLLM to ensure the complete preservation of event information. In addition, to ensure the stability and reproduction of our results, we adopt the greedy search as the default decoding strategy for all open-source models. For the closed-source models, we sample 8 frames.

\begin{table}[t]
\centering
\scalebox{0.85}{
\begin{tabular}{@{}lcccc@{}}
\toprule
Method         & Entire & Mix & Misleading & Overall \\ \midrule
Baseline & 49.12 & 70.47 & 71.57 & 64.79 \\
TCD            & 50.88  & \textbf{73.58}      & \textbf{74.50}      & 67.48   \\
TCD+noise & \textbf{51.75}  & \textbf{73.58}      & \textbf{74.50}      & \textbf{67.73}   \\
\bottomrule
\end{tabular}}
\caption{Ablation study of different strategies when constructing the contrastive counterpart of the original video. The best results of each category are \textbf{bolded}.}
\label{tab:counterpart}
\vspace{-0mm}
\end{table}

\noindent\textbf{Performance Analysis.}\quad We exhibit the performance of each VideoLLM in Tab \ref{tab:main_table} on the binary classification and detailed description matching task. As the results indicate, open-source VideoLLMs exhibit a consistent weakness in event hallucinations. In contrast, closed-source VideoLLMs demonstrate greater robustness against event hallucinations across all categories. A deeper analysis is provided below:
(1) In evaluating the susceptibility of VideoLLMs to vision-language correlation biases (i.e., ``Entire" and ``Mix" in Tab.\ref{tab:main_table}), most VideoLLMs perform worse when confronted with videos containing a mix of common and rare events, compared to those containing only rare events. This result indicates that frequent events can distract the model's attention, imposing higher challenges to the model. 
% It is worth noting that since the binary question's content of ``Entire" and ``Mix" are different as discussed in Sec.\ref{sec:data_curation}, their results are not comparable to each other.
(2) In evaluating the susceptibility of VideoLLMs to language priors (i.e., ``Misleading" in Tab.\ref{tab:main_table}), closed-source models demonstrate overwhelming advantages compared to open-source models, suggesting that these closed-source VideoLLMs are built upon language foundation models that are significantly more powerful than their open-source counterparts.

% performance on the detailed description matching of \textit{Mix} is much lower than the result on the binary classification task, which suggests that VideoLLMs may follow the question as the cue to search rare events. However, they are prone to overlook rare events in their detailed descriptions when no cues are provided. Meanwhile, VideoLLMs have difficulty in processing both binary classification and detailed description matching on \textit{Mix}. This indicates the over-reliance of VideoLLMs on language priors. Notably, VideoLLMs show worse capacity on the description matching of \textit{Mix} than \textbf{Entire}, which proves that common events can strongly induce VideoLLMs to generate hallucinations than with only rare events. In \textit{Misleading}, despite their high accuracy on description matching, results on binary classification indicate that VideoLLMs can be affected by questions and generate event hallucinations.

% \begin{figure}[ht]
%     \centering
%     \includegraphics[width=0.48\textwidth]{figs/exp2.pdf}
%     \caption{Accuracy of TCD on binary classification of EventHallusion with different numbers of sampled frames in the distorted video. \textbf{Base} indicates the results of standard decoding with 32 frames sampled.}
%     \label{fig:exp2}
% \end{figure}

\subsection{TCD Evaluation}
\subsubsection{VideoLLM Baselines.}
To evaluate the effectiveness of our TCD in addressing event hallucinations, we adopt LLaVA-NeXT-Video \cite{li2024llava}, VideoChat2 \cite{li2024mvbench}, and VILA \cite{lin2024vila} as the foundation backbones. LLaVA-NeXT-Video extends the ability of LLaVA \cite{liu2024visual} based on the projection layer design, while VideoChat2 adopts the Q-Former \cite{li2023blip} to bridge the gap between videos and LLM. VILA, on the other hand, employs enhanced visual-language pre-training for better video comprehension. Based on these MLLMs, we explore the performance of TCD across different VideoLLM architectures and designs on EventHallusion.

\begin{figure}[t]
    \centering
    \includegraphics[width=1.0\linewidth]{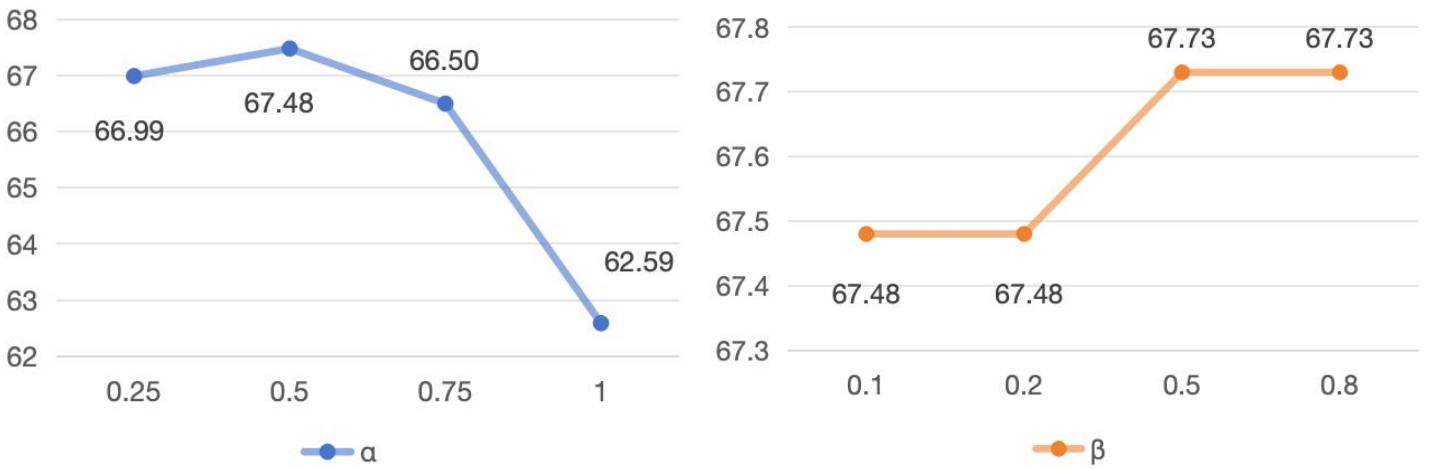}
    \caption{\textbf{Effects of $\alpha$ and $\beta$ in our TCD.}}
    \label{fig:a_b}
    \vspace{-1mm}
\end{figure}

\subsubsection{Implementation Details.}
We adopt different numbers of extracted frames to construct the original and distorted visual input: 32 and 8 frames for LLaVA-NeXT-Video, 12 and 8 frames for VILA, and 16 and 4 frames for VideoChat2. We use greedy search as the basic decoding paradigm.
% The "Default" label represents the standard decoding method using the greedy search strategy, while the "TCD" denotes our proposed Temporal Contrastive Decoding, also employing the greedy search strategy to ensure the stability of results.

% \begin{table}[ht]
% \centering
% \caption{Ablation study of extracted video frames when constructing the constrastive video. The best results of each category are \textbf{bolded}.}
% \scalebox{0.85}{
% \begin{tabular}{@{}ccccc@{}}
% \toprule
% Frames & Entire & Mix   & Misleading & Overall \\ \midrule
% 1      & 48.26  & 68.39 & 74.51      & 64.30   \\
% 4      & 50.00  & 73.06 & \textbf{76.47}      & \textbf{67.48}   \\
%  8      & \textbf{50.88}  & \textbf{73.58} & 74.51      & \textbf{67.48}   \\
% 16     & \textbf{50.88}  & 73.06 & 74.51      & 66.75   \\ \bottomrule
% \end{tabular}}
% \label{tab:fram_num}
% \end{table}

\subsubsection{Effectiveness of TCD}
We apply our proposed TCD method on three prevalent VideoLLMs, namely LLaVA-NeXT~\cite{li2024llava}, VideoChat2~\cite{li2024mvbench}, and VILA~\cite{lin2024vila}, and evaluate their performances on EventHallusion.
As shown in Tab.\ref{tab:tcd}, equipping with TCD, three VideoLLMs achieve a consistent performance boost. This demonstrates that TCD can be seamlessly integrated with versatile VideoLLMs and reduce their event-related hallucinations without bells and whistles. 
% % (+3.10 on LLaVA-NeXT-Video, +1.94 on VideoChat2, and +3.58 on VILA respectively), 
% confirming the effectiveness of TCD in mitigating event hallucinations on VideoLLMs with different architectures and visual-language pertaining paradigms. However, while TCD shows consistent and notable improvements on \textit{Mix} and \textit{Misleading} across all three models, the improvement on \textit{Entire} is minimal, with even a decrease situation observed.
% % (-2.63\% on VILA)
% This result suggests that entire rare events are a more challenging and hallucination-prone scenario for VideoLLMs.

\subsubsection{Explorations on Contrastive Sample Constructions}
We further explore more alternative strategies to construct the contrastive counterpart in our TCD method.
Recalling that our TCD aims at ambiguating the temporal cues of the original video, we further explore the effects of blurring the spatial cues by adding gaussian noises to video frames when constructing the contrastive counterpart. As shown in Tab.\ref{tab:counterpart}, introducing Gaussian noises on top of TCD can further boost the overall performance. However, the extent of introducing Gaussian noise should be deliberately adjusted to prevent degradation in practice. Therefore, we do not choose to apply this strategy in our TCD method.

% Specifically, we compare our base TCD method with adding Gaussian noises to video frames when employing TCD method (denoted as TCD+noises). As results have shown in Tab.\ref{tab:counterpart}, While both TCD and TCD+noises demonstrate improvement on regular decoding. However, TCD+noises does not exhibit large improvements on simply using TCD method.

% \usepackage{booktabs}
% \begin{table}[ht]
% \centering
% \caption{\textbf{Ablation study of $\alpha$ in our TCD.} The best results of each category are \textbf{bolded}.}
% \scalebox{0.9}{
% \begin{tabular}{@{}lcccc@{}}
% \toprule
% $\alpha$ & Entire & Mix   & Misleading & Overall \\ \midrule
% 0.25                  & 50.00  & \textbf{73.56} & 73.53      & 66.99   \\
% 0.5                   & \textbf{50.88}  & \textbf{73.56} & \textbf{74.51}      & \textbf{67.48}   \\
% 0.75                  & \textbf{50.88}  & 71.50 & \textbf{74.51}      & 66.50   \\
% 1.0                   & 49.12  & 64.25 & \textbf{74.51}      & 62.59   \\ \bottomrule
% \end{tabular}}
% \label{tab:alpha}
% \end{table}

\begin{table}[t]
\centering
\scalebox{0.85}{
\begin{tabular}{@{}ccccc@{}}
\toprule
Frames & Entire & Mix   & Misleading & Overall \\ \midrule
1      & 48.26  & 68.39 & 74.51      & 64.30   \\
4      & 50.00  & 73.06 & \textbf{76.47}      & \textbf{67.48}   \\
8      & \textbf{50.88}  & \textbf{73.58} & 74.51      & \textbf{67.48}   \\
16     & \textbf{50.88}  & 73.06 & 74.51      & 66.75   \\ \bottomrule
\end{tabular}}
\caption{Ablation study of extracted video frames when constructing the contrastive video.}
\label{tab:fram_num}
\vspace{-2mm}
\end{table}

% \begin{table}[ht]
% \centering
% \caption{\textbf{Ablation study of $\beta$ in our TCD.} The best results of each category are \textbf{bolded}.}
% \scalebox{0.9}{
% \begin{tabular}{@{}lcccc@{}}
% \toprule
% $\beta$  & Entire & Mix   & Misleading & Overall \\ \midrule
% 0.1   & \textbf{50.88}  & 73.58 & \textbf{74.51}      & 67.48   \\
% 0.2   & \textbf{50.88}  & 73.58 & \textbf{74.51}      & 67.48   \\
% 0.5   & \textbf{50.88}  & 74.09 & \textbf{74.51}       & \textbf{67.73}   \\ 
% 0.8   & 50.00      & \textbf{74.61} &    \textbf{74.51}    &   \textbf{67.73}\\\bottomrule
% \end{tabular}}
% \label{tab:beta}
% \end{table}

\subsubsection{Ablations on Hyper-parameters of TCD}
We also ablate the effects of $\alpha$ and $\beta$ in our TCD method. As shown in Fig.\ref{fig:a_b}, we have the following observations: (1) With $\alpha$ increases from 0.25 to 0.5, TCD enjoys the benefits of the contrastive video's modulation. 
However, further increasing $\alpha$ to 1.0 causes TCD to degrade significantly, suggesting that excessively amplifying the influence of original samples 
% can lead the model to deviate from its initially correct predictions.
will cause the model to decode primarily according to the original distribution, rendering TCD ineffective.
% But further increasing the $\alpha$ to 1.0, TCD degrades drastically, which indicates that excessively amplifying the effects of contrastive samples can deviate initial correct prediction of the model. We believe this phenomenon arises from the increase in $\alpha$, which excessively amplifies the significance of logits in the original video input while diminishing the reference to the distorted counterpart during the contrastive decoding process. Consequently, the model becomes overly reliant on the original output distributions, leading to the failure of TCD.
(2) Increasing $\beta$ from 0.2 to 0.5 enhances the model's performance, while further decreasing or increasing $\beta$ does not result in performance fluctuations. This demonstrates that our TCD is relatively robust to the variation in $\beta$ values.

% \begin{figure}[t]
%     \centering
%     \includegraphics[width=0.79\linewidth]{figs/tcd_example2.pdf}
%     \caption{Qualitative Results of the TCD method on our EventHallusion Benchmark.}
%     \label{fig:exam}
%     \vspace{-3mm}
% \end{figure}

\subsubsection{Effect of Sampled Frames in Distorted Video.}
We further explore the effects of the temporal downsampling ratio when constructing the contrastive video in our TCD. Specifically, we adopt LLaVA-NeXT-Video\cite{li2024llava} as the backbone model with 32 frames sampled for the original video input and downsample the number of sampled frames to construct the contrastive video. This experiment is conducted on the binary classification task of EventHallusion. From Tab.\ref{tab:fram_num}, we have the following observations: (1) Extreme temporal downsampling (i.e., 1 frame) can lead to severe performance degradation of our TCD approach. This phenomenon may arise because single-frame input significantly diminishes spatial cues, leading to distorted prediction logits during contrastive learning. (2) A low temporal downsampling rate (e.g., 16 frames) performs worse than higher rates (e.g., 4 and 8 frames). This indicates that narrowing the gap between the contrastive video and the original one weakens the advantages of our TCD approach.
% TCD consistently demonstrates steady performance improvements when the number of sampled frames ranges from 4 to 16. This indicates that  However, when the number of sampled frames decreases from 4 to 2, the performance of TCD deteriorates. We attribute this phenomenon to the excessive destruction of the temporal modeling in the VideoLLM with too few frames incorporated. This leads to drastic changes in the logit of distorted video, thereby affecting the contrastive decoding performance.

% \subsection{Qualitative Results}
% We also present qualitative results comparing the baseline method with and without our TCD approach on the EventHallusion benchmark. As shown in Fig.\ref{fig:exam}, The baseline method fails to generate accurate responses, whereas incorporating our TCD approach yields correct answers. Due to the page limitation, we provide more examples in our supplementary materials.

\section{Conclusion}
In this paper, we present EventHallusion, a novel benchmark designed to systematically evaluate the event-related hallucination issues in existing VideoLLMs. To address these hallucinations, we propose a simple yet effective Temporal Contrastive Decoding (TCD) method, which can be seamlessly integrated into prevalent VideoLLMs in a plug-and-play fashion. Extensive experimental results demonstrate the challenging nature of EventHallusion and the effectiveness of the proposed TCD approach.
% of existing VideoLLMs in a lightweight plug-and-play manner. 
% We hope our research can benefit further research in the field of reducing the hallucinations of VideoLLMs.

\bibliography{aaai2026}

\clearpage
\appendix

\section{Comparison with Other Video Hallucination Benchmarks.}
\begin{figure*}[hb]
    \centering
    \includegraphics[width=1.\linewidth]{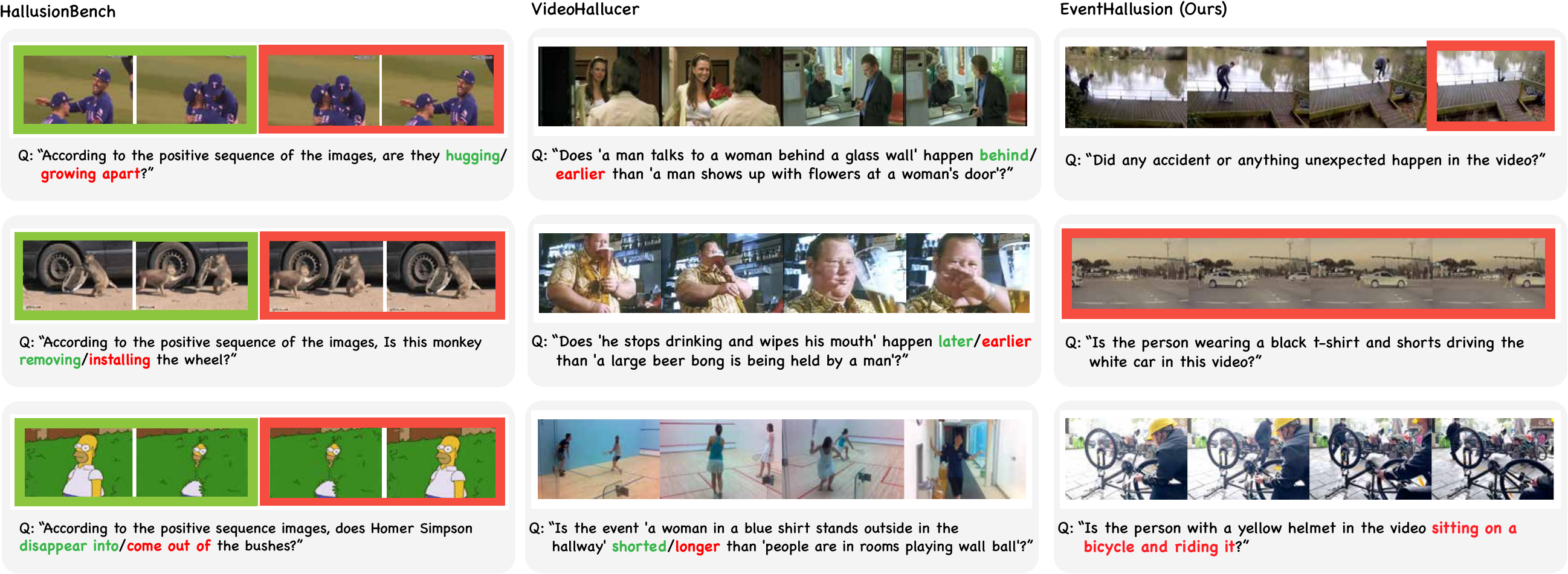}
    \caption{Comparison between HallusionBench, VideoHallucer, and our EventHallusion. We use green to highlight the content of the original video/question, and red to denote refactored video/question content to assess the model's event-related hallucinations. }
    \label{fig:compare}
\end{figure*}

We exhibit the comparison between HallusionBench, VideoHallucer, and our EventHallusion in Figure \ref{fig:compare}. As shown in Figure \ref{fig:compare}, compared to HallusionBench and VideoHallucer, which emphasize assessing the models' sensitivity to \textbf{\textit{events' chronological orders}}, our proposed EventHallusion focuses on analyzing the \textbf{\textit{event-related language prior}} and \textbf{\textit{vision-language correlation bias}} encapsulated in existing VideoLLMs.

\section{Automatic Evaluation for Detailed Description Matching}
\begin{figure*}[hb]
    \centering
    \includegraphics[width=1.\linewidth]{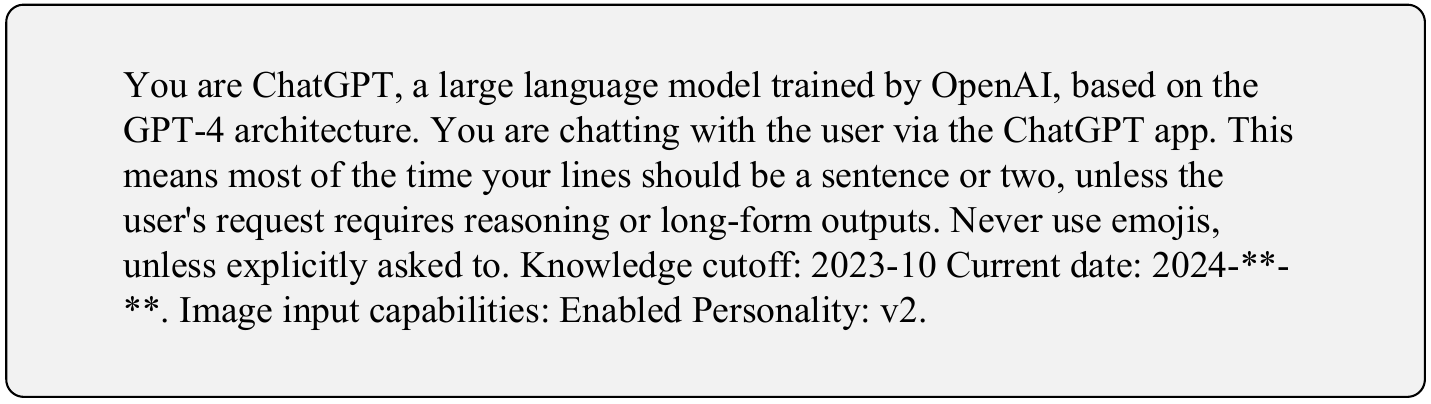}
    \caption{System prompt used in all automatic evaluation.}
    \label{fig:system}
\end{figure*}

\begin{figure*}
    \centering
    \includegraphics[width=1.\linewidth]{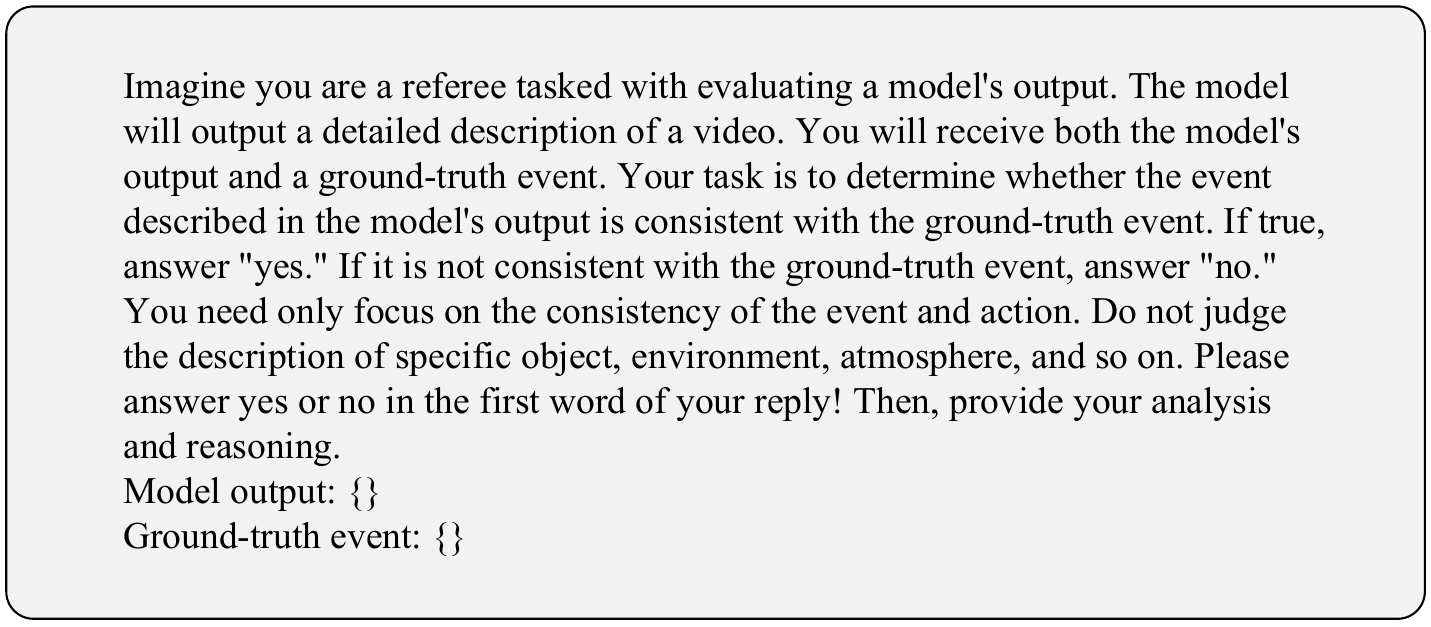}
    \caption{Prompt used for detailed description matching in \textit{Entire}.}
    \label{fig:rare_prompt}
\end{figure*}

\begin{figure*}
    \centering
    \includegraphics[width=1.\linewidth]{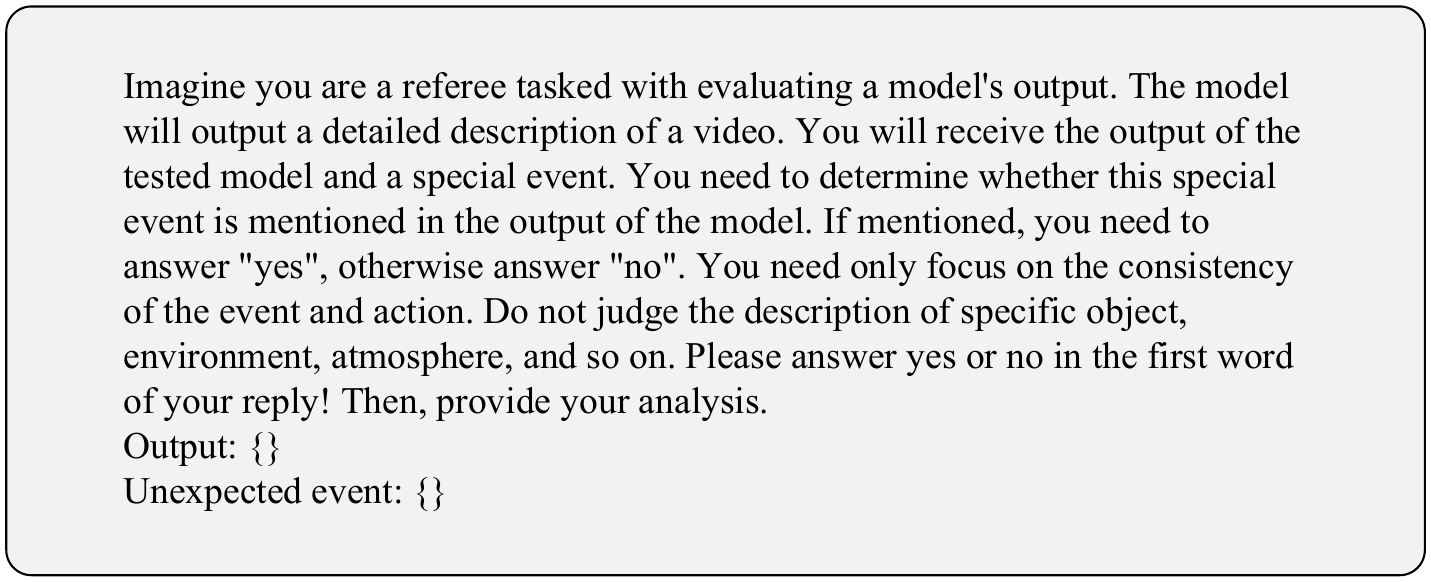}
    \caption{Prompt used for detailed description matching in \textit{Mix}.}
    \label{fig:unex_prompt}
\end{figure*}

\begin{figure*}
    \centering
    \includegraphics[width=1.\linewidth]{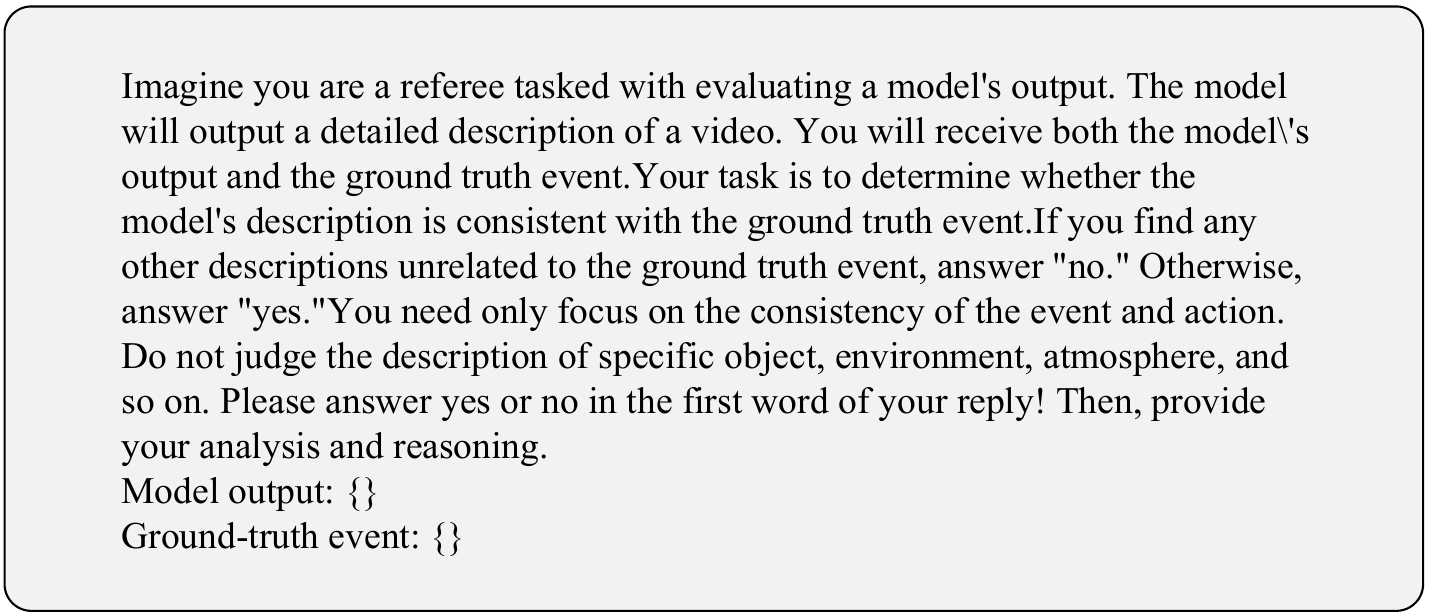}
    \caption{Prompt used for detailed description matching in \textit{Misleading}.}
    \label{fig:normal_prompt}
\end{figure*}

We apply GPT-4o to evaluate the detailed description matching task of our \textit{Entire}, \textit{Mix}, and \textit{Misleading} categories, respectively. For \textit{Entire}, we prompt GPT-4o to judge if VideoLLMs' detailed description of the video is consistent with the ground-truth rare event. And for \textit{Mix}, we prompt GPT4o to focus on the occurrence of the rare-common mixed event. Finally, for \textit{Misleading}, we prompt GPT4o to assess the consistency of the ground-truth event and search for any unrelated events. In addition, we set up the default system prompt to better guide GPT-4o's evaluation, which is shown in Figure \ref{fig:system}. The detailed prompts are shown in Figure ~\ref{fig:rare_prompt}, \ref{fig:unex_prompt}, \ref{fig:normal_prompt}, respectively.

\section{Qualitative Examples}
\begin{figure*}
    \centering
    \includegraphics[width=1.\linewidth]{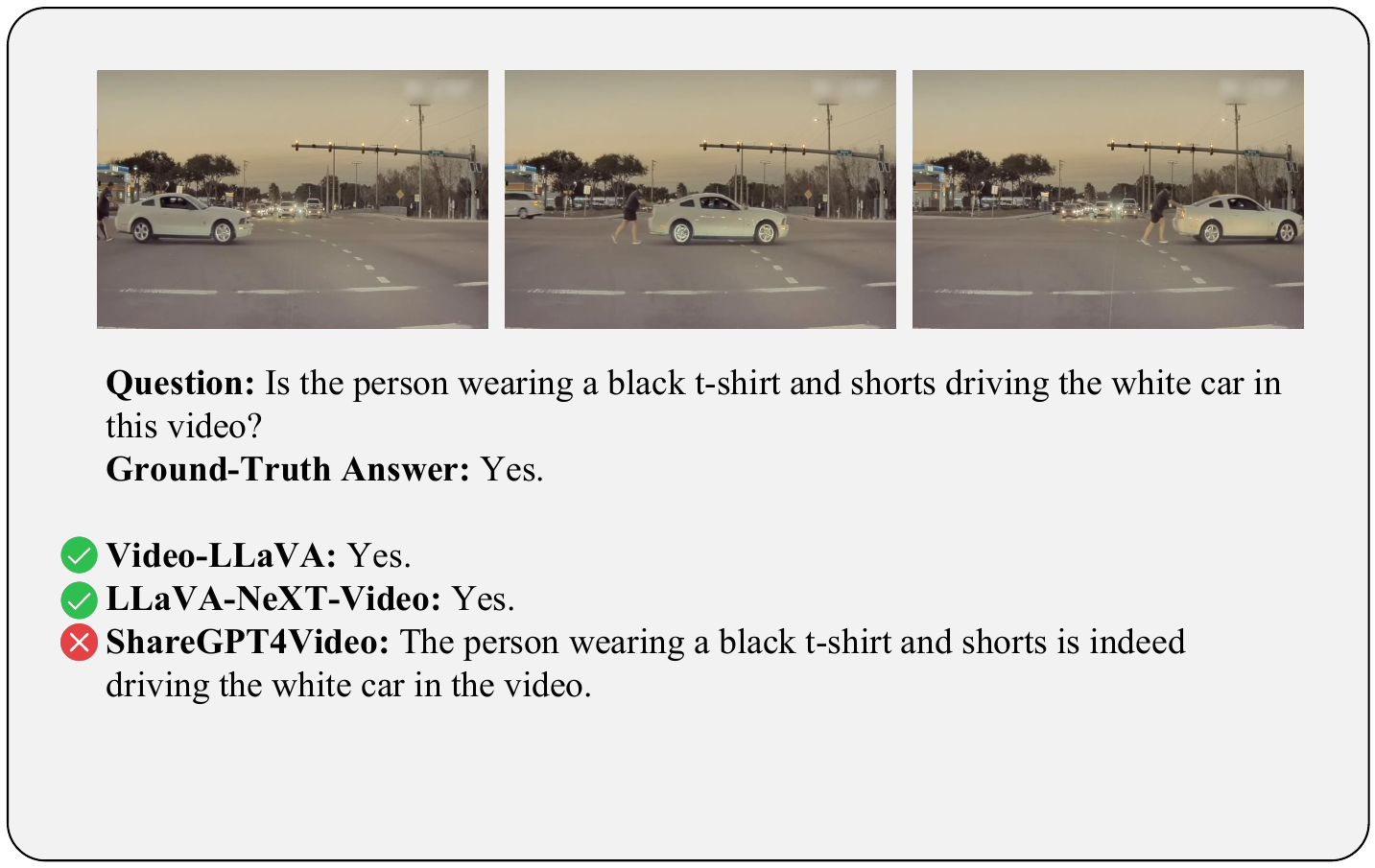}
    \caption{Examples of \textit{Entire} with responses of VideoLLMs.}
    \label{fig:rare}
\end{figure*}

\begin{figure*}
    \centering
    \includegraphics[width=1.\linewidth]{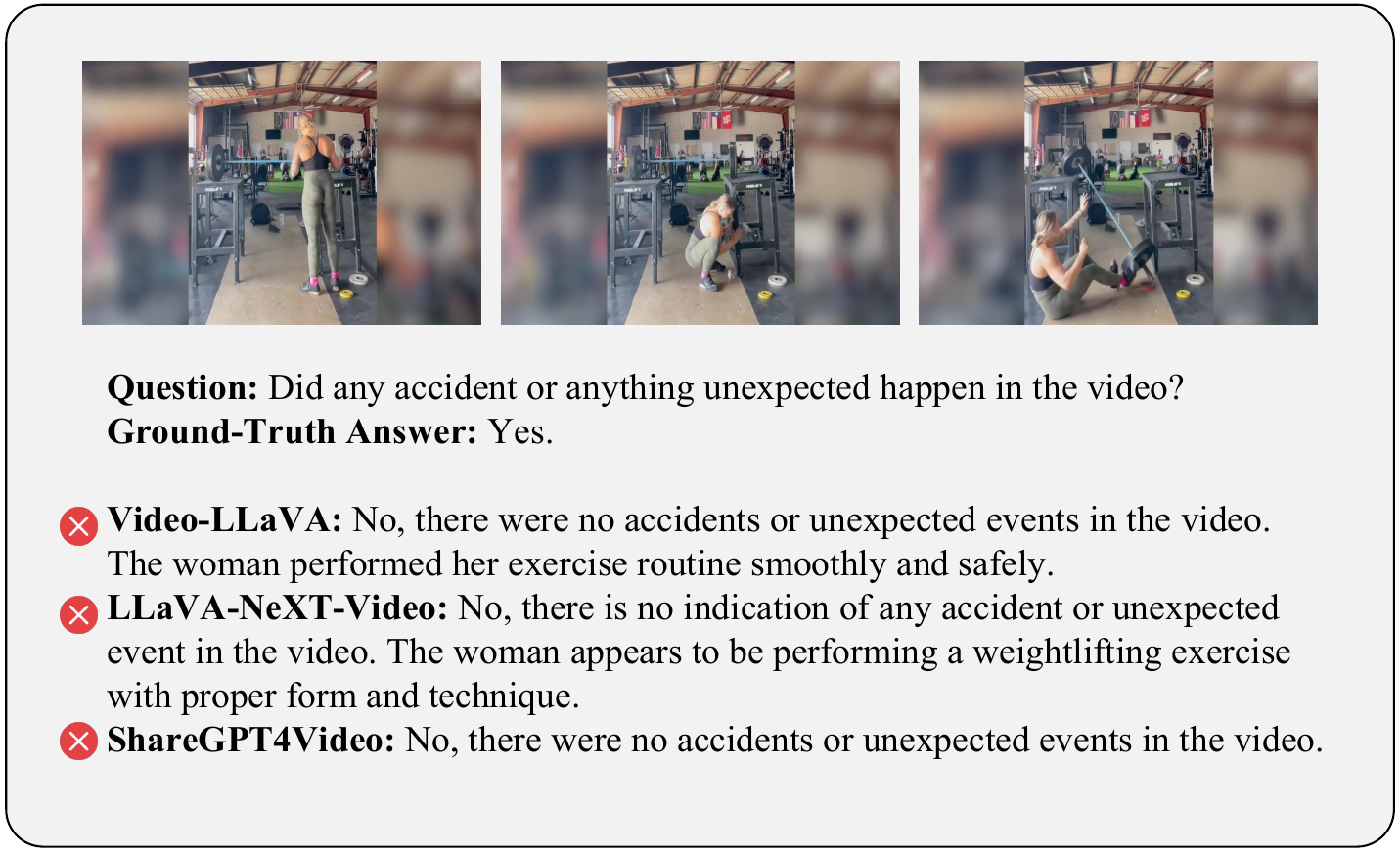}
    \caption{Examples of \textit{Mix} with responses of VideoLLMs.}
    \label{fig:unex}
\end{figure*}

\begin{figure*}
    \centering
    \includegraphics[width=1.\linewidth]{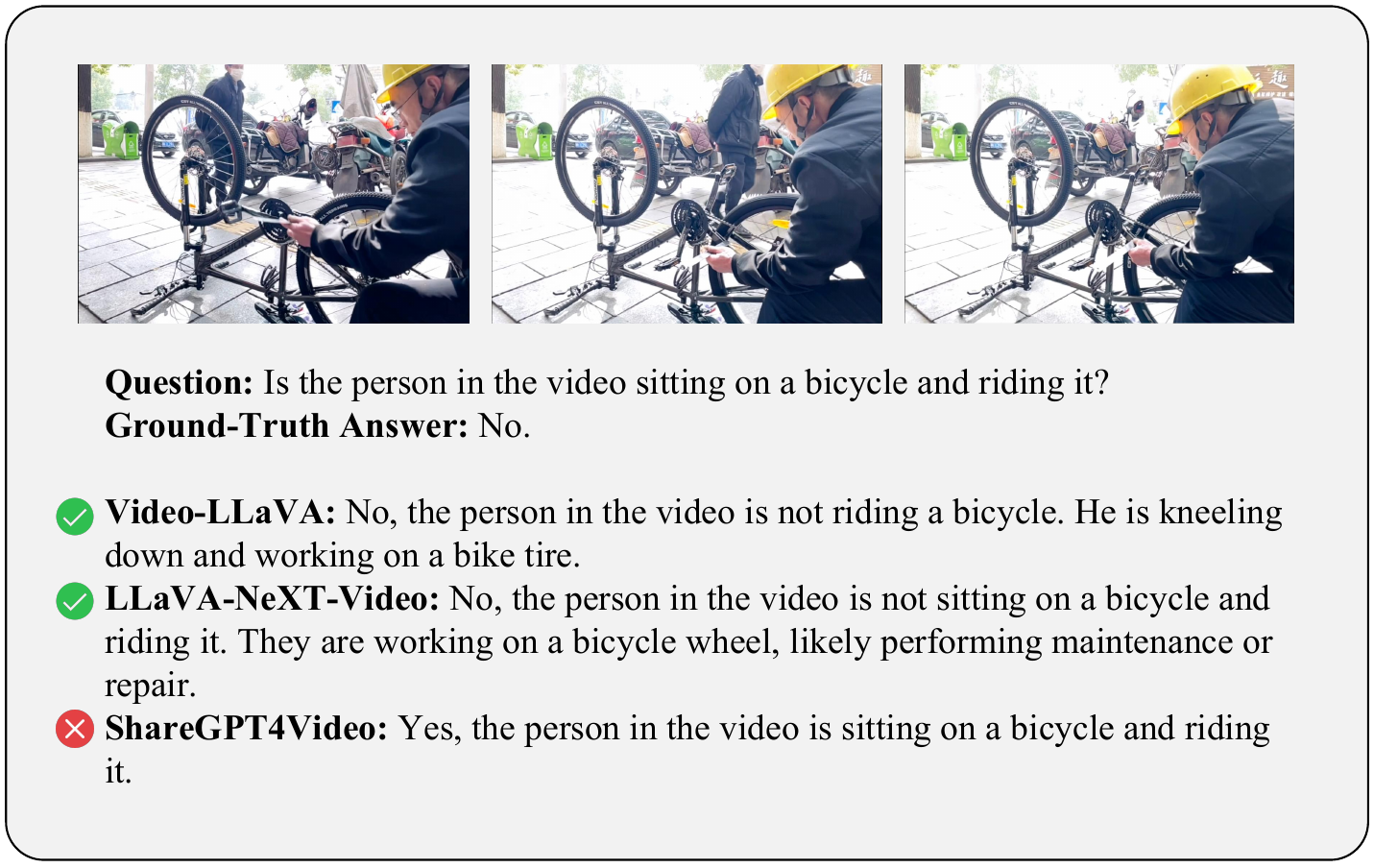}
    \caption{Examples of \textit{Misleading} with responses of VideoLLMs.}
    \label{fig:normal}
\end{figure*}

\noindent{\textbf{Data examples of EventHallusion.}}
We present examples of \textit{Entire}, \textit{Mix}, and \textit{Misleading} in Figure \ref{fig:rare}, \ref{fig:unex}, \ref{fig:normal}, incorporating responses of VideoLLMs. As shown in these examples, VideoLLMs are prone to generate hallucinated answers under all of our settings, indicating their deficiency in addressing event hallucinations.

\begin{figure*}
    \centering
    \includegraphics[width=1.\linewidth]{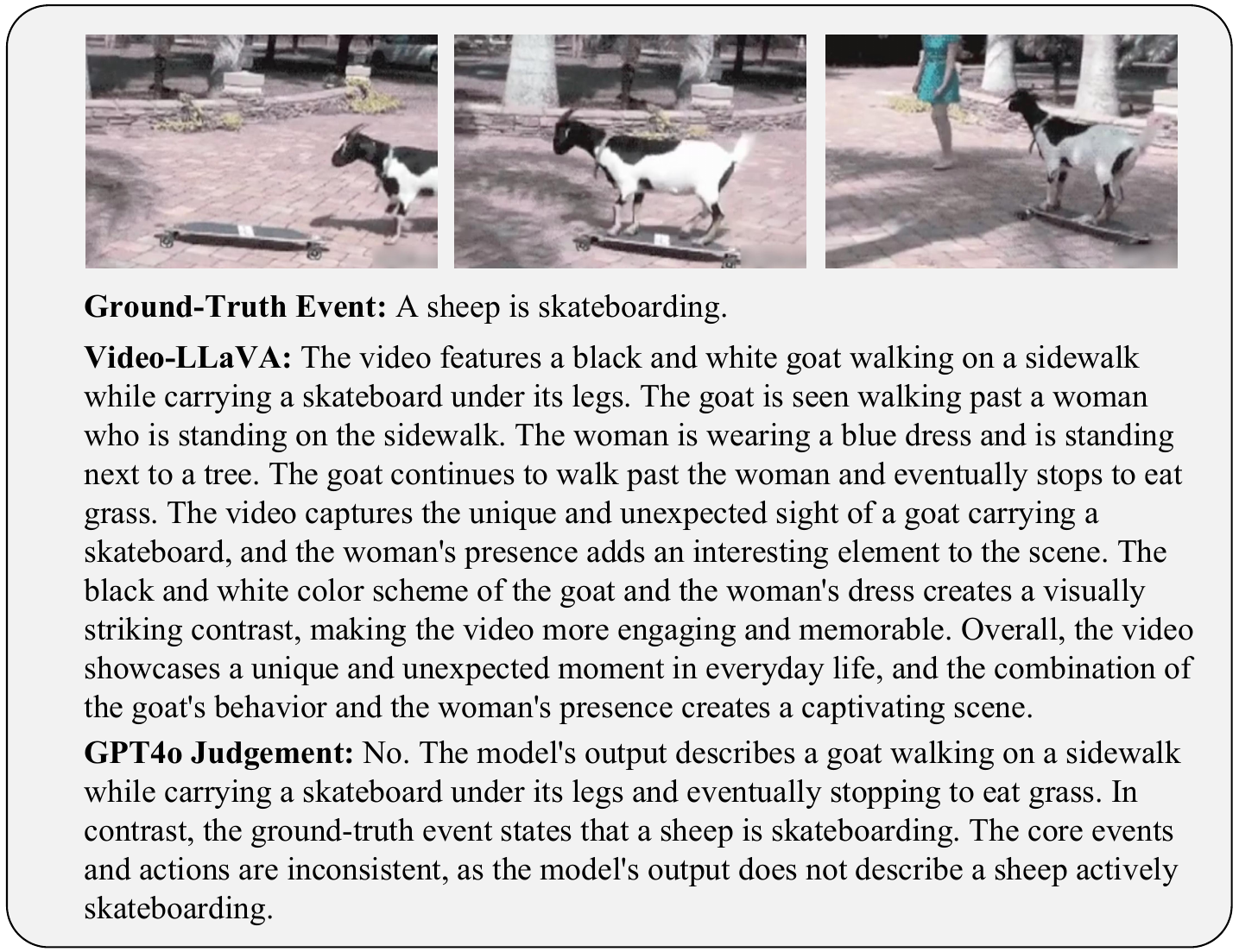}
    \caption{Examples of the evaluation of GPT-4o on the detailed description of VideoLLMs of \textit{Entire}.}
    \label{fig:rare_gpt}
\end{figure*}

\begin{figure*}
    \centering
    \includegraphics[width=1.\linewidth]{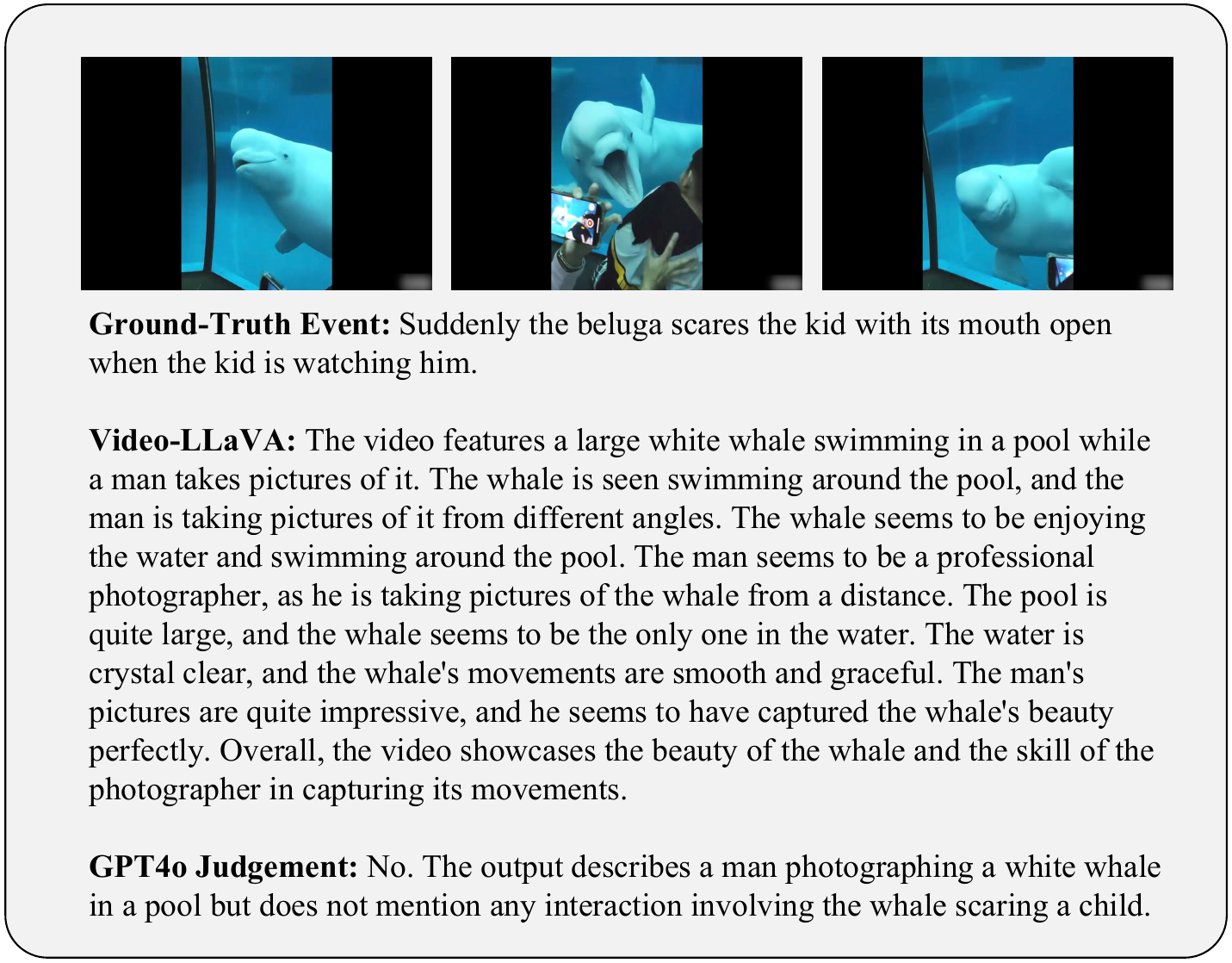}
    \caption{Examples of the evaluation of GPT-4o on the detailed description of VideoLLMs of \textit{Mix}.}
    \label{fig:unex_gpt}
\end{figure*}

\begin{figure*}
    \centering
    \includegraphics[width=1.\linewidth]{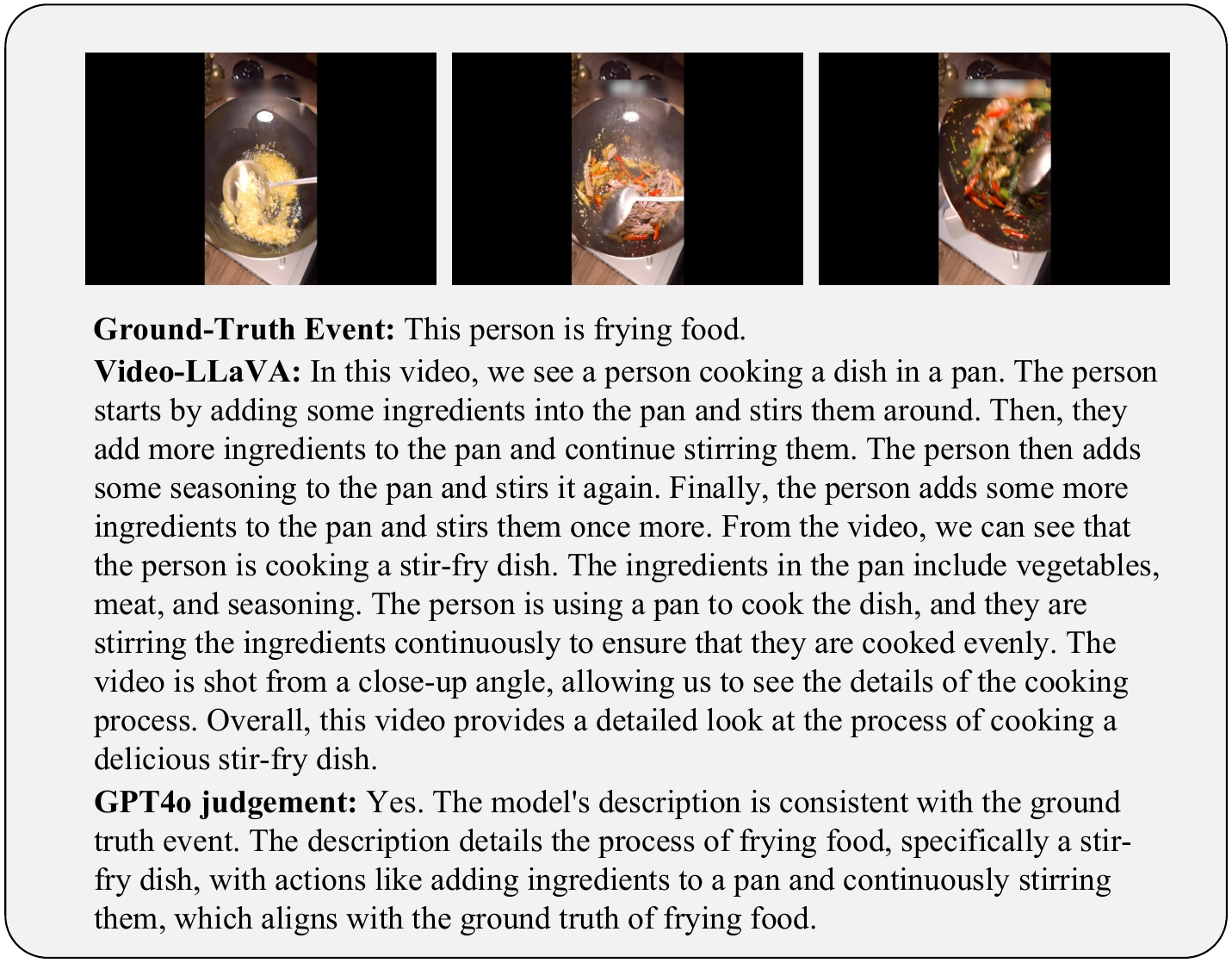}
    \caption{Examples of the evaluation of GPT-4o on the detailed description of VideoLLMs of \textit{Misleading}.}
    \label{fig:normal_gpt}
\end{figure*}

\noindent{\textbf{Examples of GPT-4o's Evaluation.}}
We present examples of GPT-4o's evaluation of detailed description matching task on \textit{Entire}, \textit{Mix}, and \textit{Misleading} respectively in Figure \ref{fig:rare_gpt}, \ref{fig:unex_gpt}, \ref{fig:normal_gpt}. As shown in these examples, GPT-4o has the capacity to judge the accuracy of video events. 

\begin{figure*}
    \centering
    \includegraphics[width=1.\linewidth]{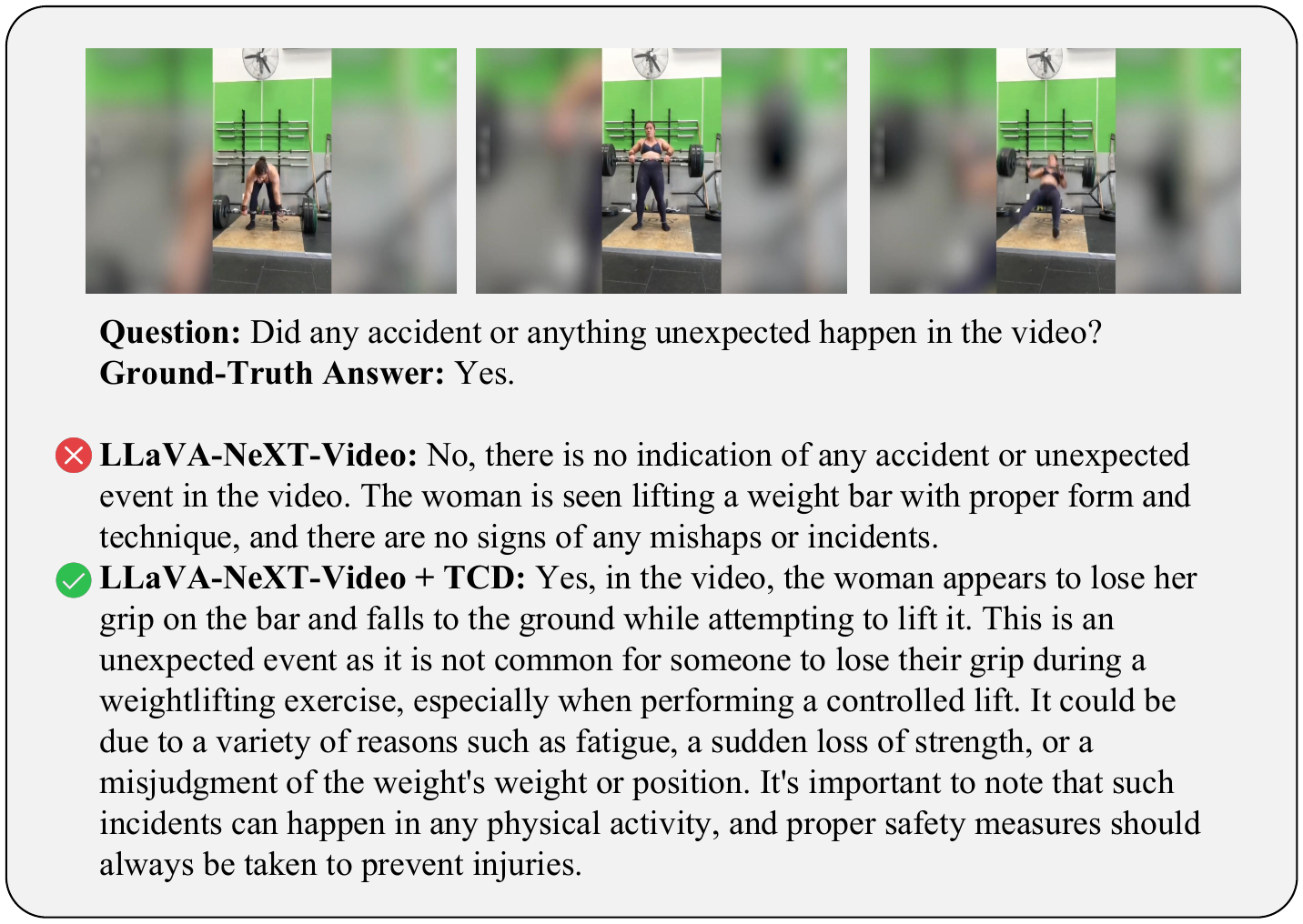}
    \caption{Examples of TCD with LLaVA-NeXT-Video on our EventHallusion.}
    \label{fig:tcd1}
\end{figure*}

\begin{figure*}
    \centering
    \includegraphics[width=1.\linewidth]{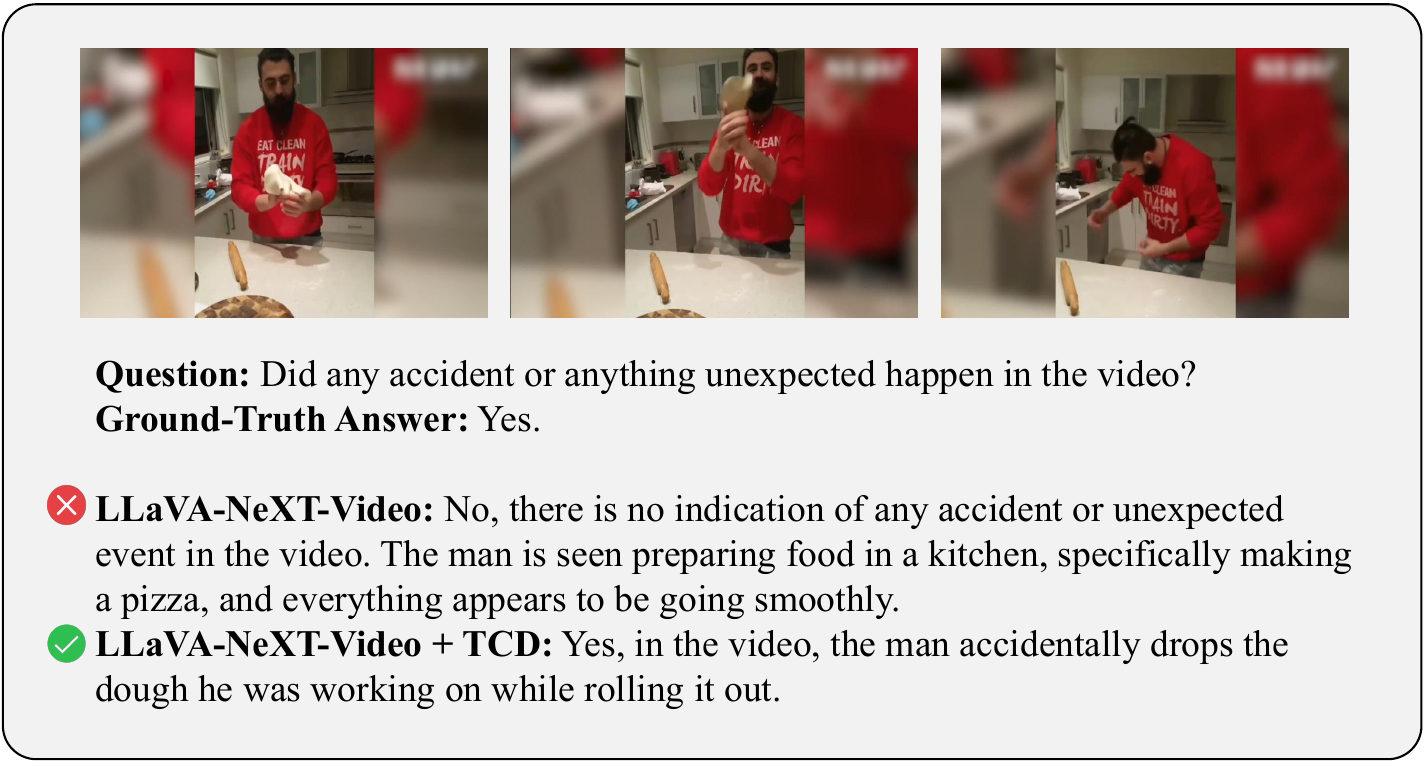}
    \caption{Examples of TCD with LLaVA-NeXT-Video on our EventHallusion.}
    \label{fig:tcd2}
\end{figure*}

\begin{figure*}
    \centering
    \includegraphics[width=1.\linewidth]{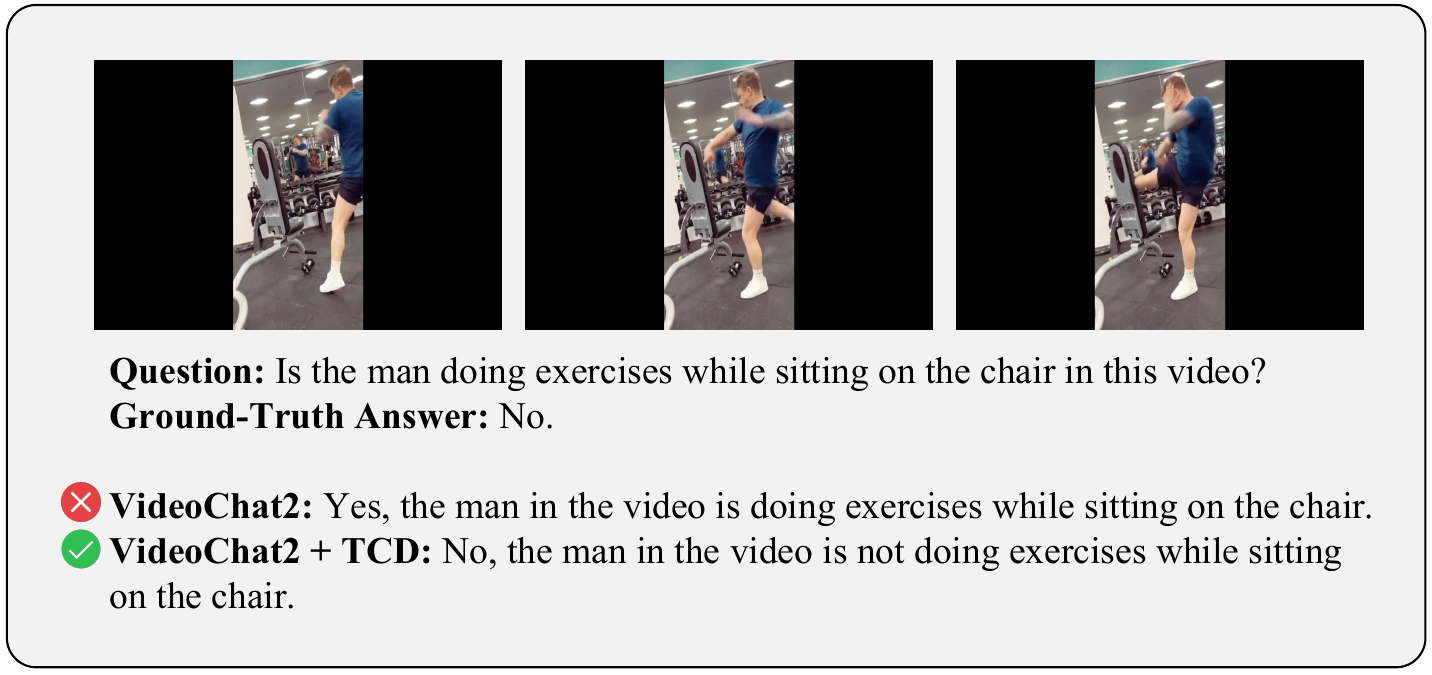}
    \caption{Examples of TCD with VideoChat2 on our EventHallusion.}
    \label{fig:tcd3}
\end{figure*}

\noindent{\textbf{Examples of TCD.}}
We present examples of our TCD in mitigating video event hallucinations. Specifically, we exhibit the generated response with and without TCD for LLaVA-NeXT-Video and VideoChat2 on our EventHallusion in Figure \ref{fig:tcd1}, \ref{fig:tcd2}, and \ref{fig:tcd3}.

% Prompt

% Data examples

% GPT-4o evaluation examples

% TCD examples

\end{document}